\providecommand{\main}{.}
\documentclass[10pt,conference]{IEEEtran}
\usepackage[left=45pt,right=45pt,top=54pt,bottom=72pt]{geometry}

\usepackage{times}
\usepackage{subfiles}
\usepackage{subfig}

\usepackage[numbers]{natbib}
\usepackage{multicol}

\usepackage[T1]{fontenc}
\usepackage[utf8]{inputenc}
\usepackage{algorithm}
\usepackage{algpseudocode}
\usepackage{amsfonts}
\usepackage{amsmath}
\usepackage{amssymb}
\usepackage{amsthm}
\usepackage{bbm}
\usepackage{bm}
\usepackage{booktabs}
\usepackage{color}
\usepackage{enumerate}
\usepackage{footmisc}
\usepackage{graphicx}
\usepackage{import}
\usepackage{mathrsfs}
\usepackage{microtype}
\usepackage{nicefrac}
\usepackage{pgfplots}
\pgfplotsset{compat=1.5}
\usepackage{psfrag}
\usepackage{scalerel}
\usepackage{tikz}
\usepackage{url}
\usepackage{multirow}
\usepackage[super]{nth}

\newcommand{\Bez}{B\'ezier }

\newcommand{\argmin}{\mathop{\rm argmin}}

\newcommand{\eg}{{\it e.g. }}

\newcommand{\ie}{{\it i.e.}}

\newcommand{\dt}{ \,\mathrm{d}t}

\newcommand{\executeiffilenewer}[3]{%
  \ifnum\pdfstrcmp%
    {\pdffilemoddate{#1}}%
    {\pdffilemoddate{#2}}%
    >0%
    {\immediate\write18{#3}}%
  \fi%
}

\usepackage[bookmarks=true]{hyperref}
\newcommand{\dmin}{{d_\mathrm{min}}}
\newcommand{\dlb}{{d_\mathrm{lb}}}
\newcommand{\dub}{{d_\mathrm{ub}}}

\algnewcommand{\IfThen}[2]{
  \State \algorithmicif\ #1\ \algorithmicthen\ #2}

\usepackage[capitalise,nameinlink,noabbrev]{cleveref}
\theoremstyle{definition}
\newtheorem{theorem}{Theorem}

\newtheorem{lemma}{Lemma}
\newtheorem{assumption}{Assumption}

\newtheorem{problem}{Problem}

\crefname{problem}{Problem}{Problems}

\theoremstyle{remark}
\newtheorem*{remark}{Remark}

\graphicspath{{figs/}{\main/figs/}}
\usepackage[font={small}]{caption}

\begin{document}

\title{Proximity Queries for Absolutely Continuous Parametric Curves}


\author{\authorblockN{Arun Lakshmanan\authorrefmark{2},
Andrew Patterson\authorrefmark{2},
Venanzio Cichella\authorrefmark{3} and
Naira Hovakimyan\authorrefmark{2}}
\authorblockA{\authorrefmark{2}University of Illinois at Urbana-Champaign,  \authorrefmark{3}University of Iowa}
\authorblockA{lakshma2@illinois.edu, appatte2@illinois.edu, venanzio-cichella@uiowa.edu, nhovakim@illinois.edu}
}

\maketitle

\begin{abstract}
  In motion planning problems for autonomous robots, such as self-driving cars, the robot must ensure that its planned path is not in close proximity to obstacles in the environment. However, the problem of evaluating the proximity is generally non-convex and serves as a significant computational bottleneck for motion planning algorithms. In this paper, we present methods for a general class of absolutely continuous parametric curves to compute: (i) the minimum separating distance, (ii) tolerance verification, and (iii) collision detection. Our methods efficiently compute bounds on obstacle proximity by bounding the curve in a convex region. This bound is based on an upper bound on the curve arc length that can be expressed in closed form for a useful class of parametric curves including curves with trigonometric or polynomial bases. We demonstrate the computational efficiency and accuracy of our approach through numerical simulations\footnote{The implementation can be found at \href{https://github.com/arlk/CurveProximityQueries.jl}{\texttt{https://github.com/arlk/\allowbreak CurveProximityQueries.jl}}.\label{repo}} of several proximity problems.
\end{abstract}

\IEEEpeerreviewmaketitle

\section{Introduction}
\label{sec:intro}
Autonomous robots often operate in rapidly changing environments, and the ability to accurately and quickly predict future collisions is crucial for safely meeting task objectives. As proximity queries serve  one of the major bottlenecks in motion planning frameworks \cite{kleinbort2016}, algorithms that efficiently assess the proximity of obstacles relative to the future states of the robot will greatly improve the run-time performance of the robot while guaranteeing safe operating procedures.

Fast methods for computing proximity queries between polyhedral objects have been widely studied and developed over the past few decades. The algorithms described in \cite{gilbert1988, lin1991} perform several different types of proximity queries between convex polytopes. Proximity between general polyhedral objects using hierarchical representations of convex bounding volumes are investigated in \cite{larsen1999} and \cite{gottschalk1996}. In \cite{ehmann2001}, the authors evaluate the proximity queries for polyhedral objects using convex surface decomposition.

It is challenging to assess the proximity between objects in motion. Recent work has focused on developing methods for computing continuous collision detection (CCD), a type of query that evaluates the first time of contact between objects in motion. The most practical of these methods is known as conservative advancement, introduced in \cite{zhang2006} and \cite{tang2009}, and performs several static proximity queries between polyhedral or polygon-soup objects to accurately perform CCD queries. Conservative advancement has been applied in trajectory refinement algorithms \cite{pan2012} to obtain collision-free cubic B-spline trajectories.  However, in the context of motion planning, such methods are computationally expensive and lack the ability to evaluate other proximity queries such as the minimum separating distance or tolerance verification between moving objects. As modern motion planning and trajectory optimization methods generate candidate paths as rectifiable parametric curves (\eg~Lagrange polynomials \cite{ross2012}, Legendre polynomials \cite{mellinger2012}, \Bez curves \cite{cichella2018, cichella2018b, ricciardi2018}, B-splines \cite{van2015}, Dubins paths \cite{vana2018}, Pythagorean Hodograph curves \cite{puig2018}), the proximity algorithms must be able to handle such representations.

Generally, proximity to parametric curves is handled by algberaic, interval analysis, or curve subdivision methods. Finding multiple intersections between parametric curves by implicitization and eigenvalue decomposition is discussed in \cite{manocha1994}. In \cite{elber2008}, the authors present an approach to compute the exact minimum separating distance between differentiably continuous freeform curves by solving a set of nonlinear equations. Computing the minimum separating distance between \Bez curves by sweeping a sphere along one of the curves and eliminating sections of the curve which lie outside of it through subdivision was introduced in \cite{chen2009}. A similar technique \cite{oh2012} was used  to find the closest point on a free form curve to a point in free space. In \cite{chang2011}, the authors present an efficient method for computing the minimum separating distances to \Bez curves using subdivision methods. The curves are recursively subdivided until the bounds on the minimum separating distance between the control polygons for each of the curves are within some prescribed accuracy. However, these methods do not hold in general for parametric curves and lack the computational efficiency to be used in motion planning algorithms.

\begin{figure}[t]
  \centering
  \subfloat[]{\includegraphics[width=0.5\columnwidth]{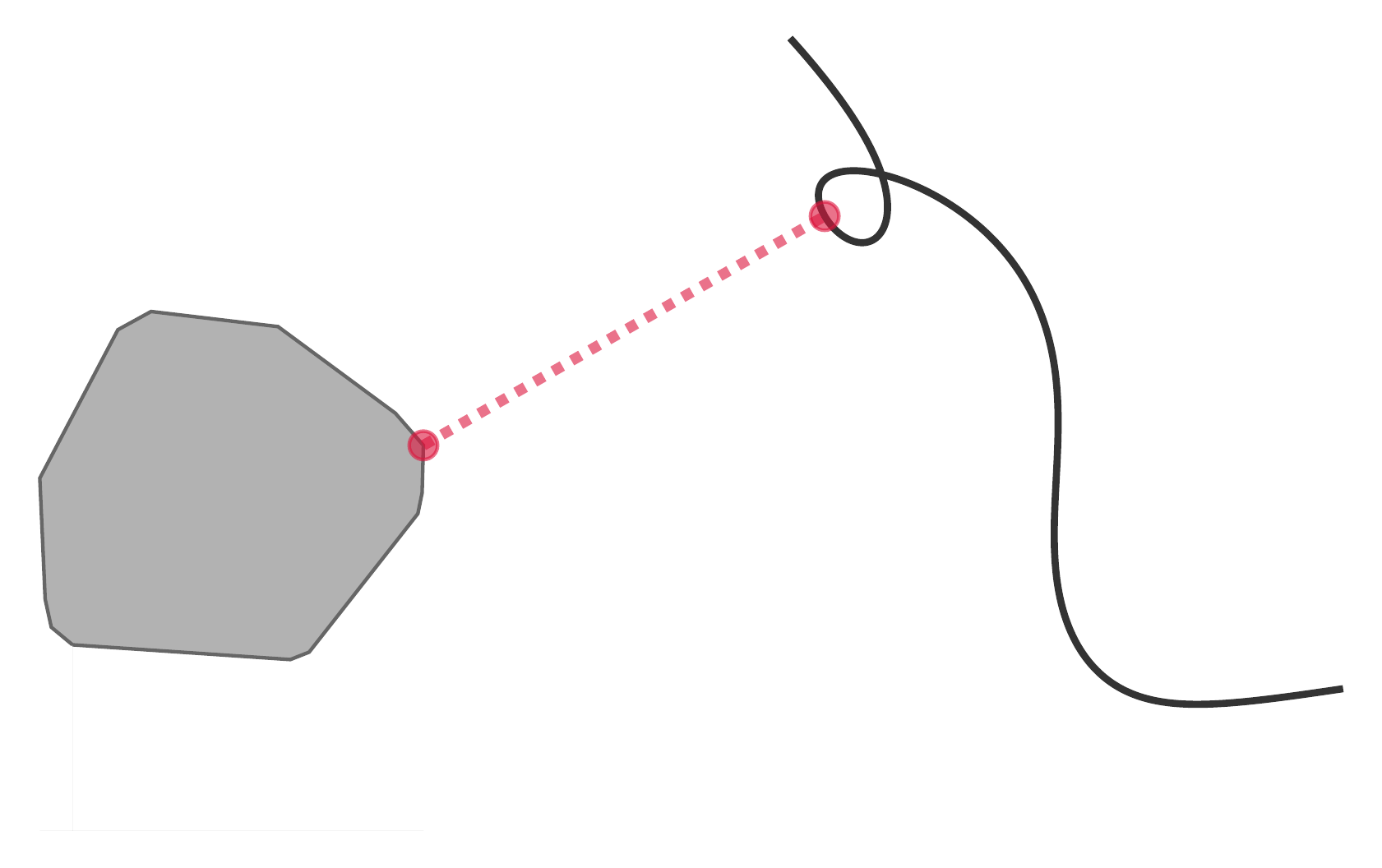}}
  \subfloat[]{\includegraphics[width=0.5\columnwidth]{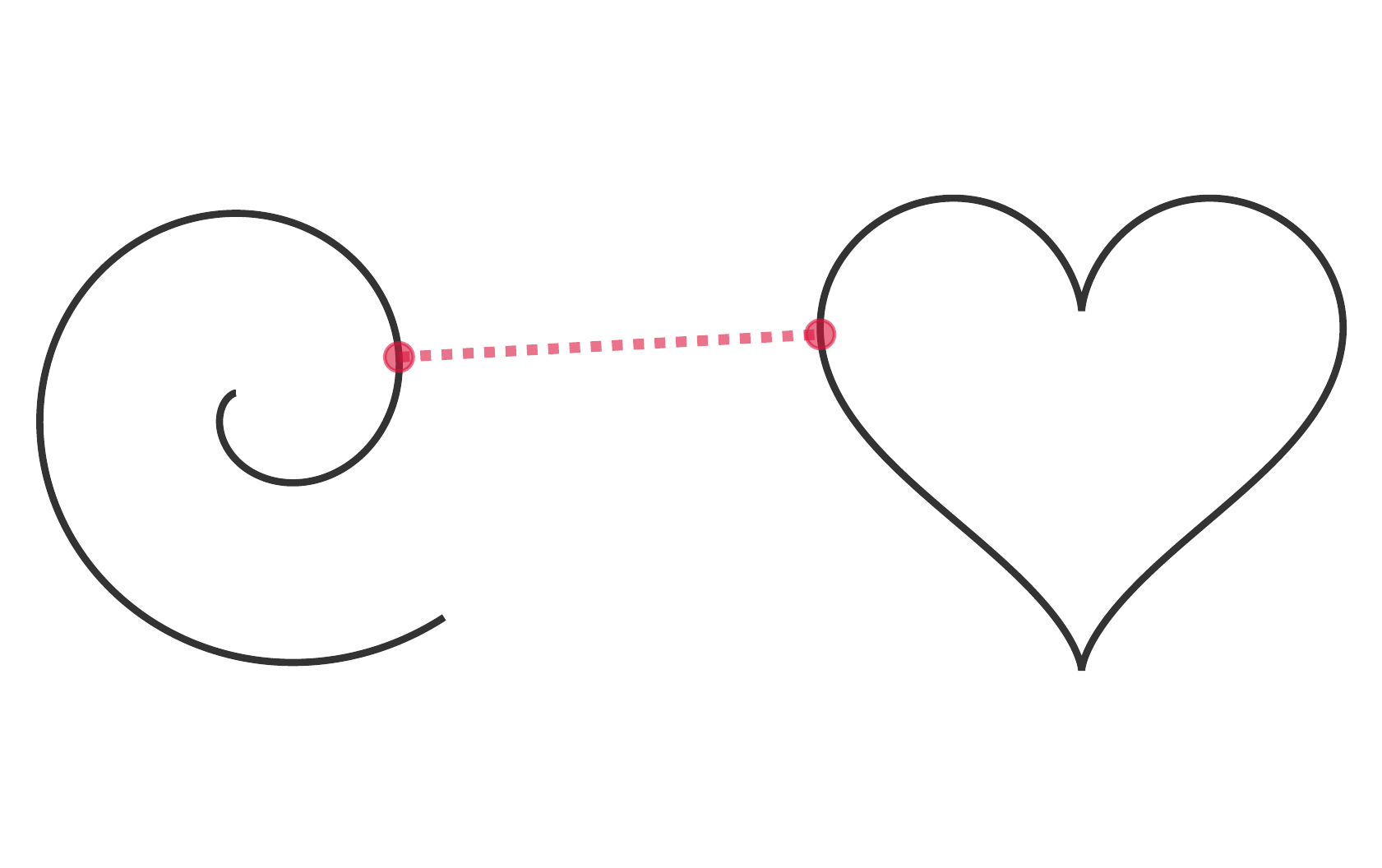}}
  \caption{Points closest between (a) a convex polygon and a \Bez curve, and (b) an involute of a circle and a heart-shaped curve.}
  \label{fig:intro}
\end{figure}

In this paper, we introduce a family of algorithms to evaluate the (i) minimum separation distance, (ii) tolerance verification, and (iii) collision detection queries between absolutely continuous parametric curves and obstacles. The obstacles are defined  as convex polytopes, parametric curves, or any other compact set to which minimum distances can be computed,  \cref{fig:intro}. A main feature of the proposed algorithms is their ability to provide proximity queries for a large class of parametric curves, more general than ones that have been considered previously \cite{manocha1994, elber2008, chen2009, oh2012, chang2011}. Such queries are useful in scenarios when the motion planner's candidate path (i) incurs a distance-based penalty for approaching close to an obstacle, (ii) must keep a safe distance from an obstacle, or (iii) must not intersect with the obstacle's geometry.

Our main contributions are summarized as follows:
\begin{itemize}
    \item Efficient computation of convex hulls for a general class of absolutely continuous parametric curves based on the arc length for any sub-interval in their domain.
    \item Fast procedures to evaluate the minimum separating distance, tolerance verification, and collision detection queries using interval branch-and-bound methods.
\end{itemize}

We present the problem formulation for the different query types in \cref{sec:intro}. The analysis and results for the construction of convex hulls for parametric curves that provide bounds on the minimum separating distance is provided in \cref{sec:bounds}. In \cref{sec:alg}, the algorithms for each of the proximity queries are outlined. Finally, in \cref{sec:results} we present numerical examples and benchmarks to illustrate the efficacy of these methods.

\section{Problem Formulation}
\label{sec:prob}
The parametric equation of a curve is given by a function $\psi : \mathcal{I} \to \mathbb{R}^d$, where $\mathcal{I} \subset \mathbb{R}$ is a compact interval such that $|\mathcal{I}| > 0$, and $d \in \mathbb{N}$. We define the curve over the closed sub-interval $\mathcal{Q} \subseteq \mathcal{I}$ (such that $|\mathcal{Q}| > 0$) as the following compact set
\begin{equation}\label{eq:curve}
    \Psi_\mathcal{Q} = \{ \psi(t) \in \mathbb{R}^d : t \in \mathcal{Q} \}.
\end{equation}
We proceed to define the proximity query problems between a curve $\Psi_\mathcal{I}$ and an object $\mathcal{B}$ represented as a nonempty compact set in $\mathbb{R}^d$ under the following assumptions.

\begin{assumption}
The function $\psi$ is absolutely continuous over its entire domain $\mathcal{I}$, \ie ,\ for every $\epsilon > 0$ there exists a $\delta > 0$, such that for each $n \in \mathbb{N}$, if the collection of mutually disjoint closed sub-intervals $\{[\alpha_i, \beta_i] \mid i = 1, \dots, n \}$ satisfies $\sum_i |\alpha_i - \beta_i | < \delta$, then $\sum_i \| \psi(\alpha_i) - \psi(\beta_i) \| < \epsilon$.
\end{assumption}

\begin{assumption}
The function $\psi$ is not a constant map over the entire domain $\mathcal{I}$, \ie ,\ $\psi(t) \neq y$ for all $t \in \mathcal{I}$ for some $y \in \mathbb{R}^d$.
\end{assumption}

\begin{figure}[t]
  \centering
  \subfloat[]{\label{fig:defn-mindist}\includegraphics[width=0.5\columnwidth]{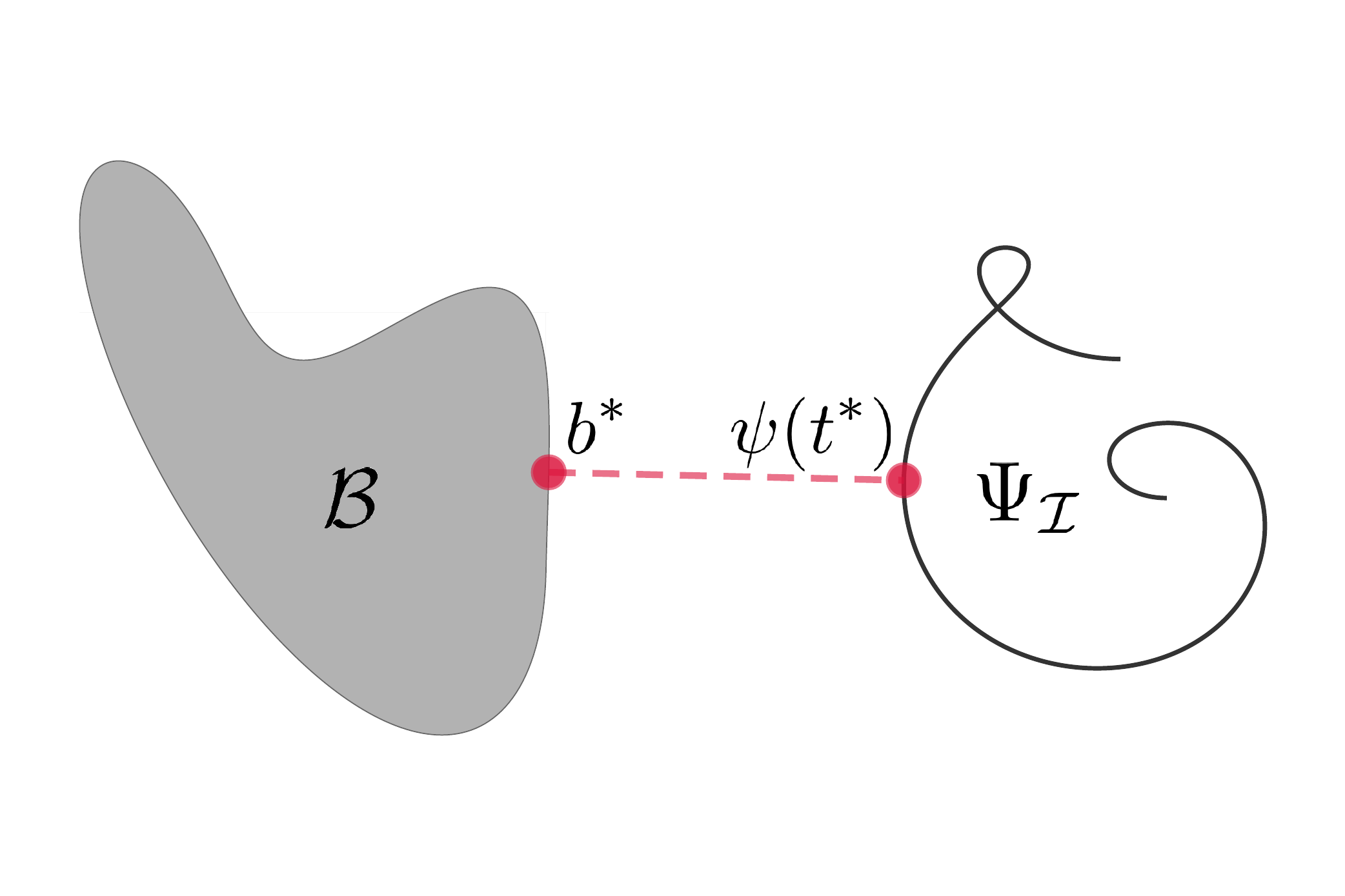}}
  \subfloat[]{\label{fig:defn-tolver}\includegraphics[width=0.5\columnwidth]{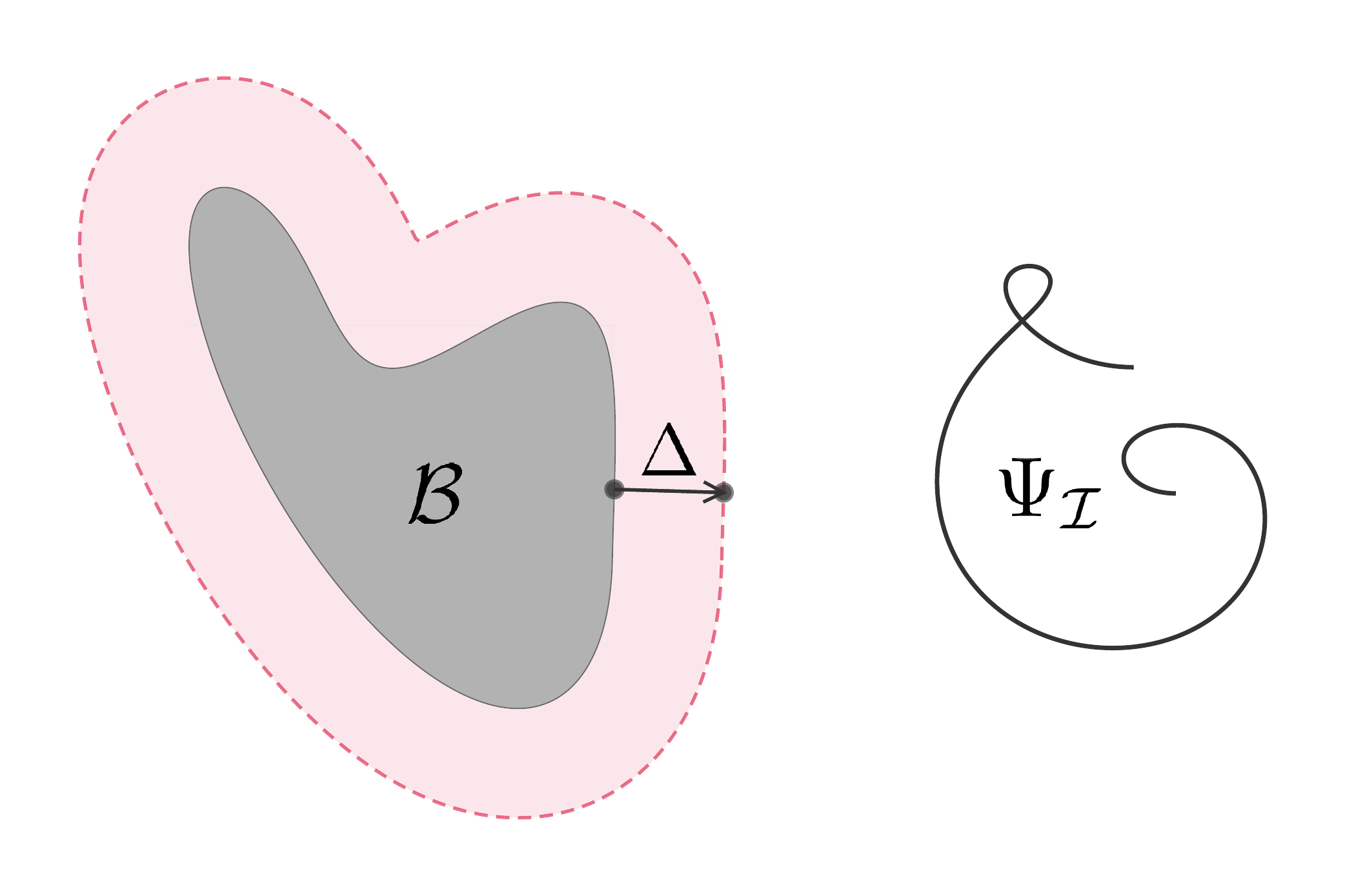}}
  \caption{(a) The red dashed line shows the segment connecting the pair of closest points between $\Psi_\mathcal{I}$ and $\mathcal{B}$ respectively. (b) The red shaded region specifies the $\Delta$-tolerance afforded to $\mathcal{B}$ when computing the tolerance verification between the two objects.}
  \label{fig:tolver}
\end{figure}

\begin{problem}[Minimum Separating Distance]
\label{problem:mindist}
The minimum separating distance between the parametric curve $\Psi_\mathcal{I}$ and the compact set $\mathcal{B}$ is defined as
\begin{equation}\label{eq:mindist}
  d_\mathrm{min}(\Psi_\mathcal{I}, \mathcal{B}) = \min_{a \in \Psi_\mathcal{I}, b \in \mathcal{B}} \| a - b \|.
\end{equation}
The points $\psi(t^*) \in \Psi_\mathcal{I}$ and $b^* \in \mathcal{B}$  that verify
\begin{equation*}
  d_\mathrm{min}(\Psi_\mathcal{I}, \mathcal{B}) = \| \psi(t^*) - b^* \|,
\end{equation*}
are the pair of points that lie closest to each other on the respective sets, as shown in \cref{fig:defn-mindist}. In addition to considering the minimum separating distance over the entire curve, $\Psi_\mathcal{I}$, the definition can be applied to some continuous compact sub-interval, $\mathcal{Q}\subseteq \mathcal{I}$, of the curve as well. This section of the curve is referred to by $\Psi_\mathcal{Q}$.
\end{problem}


\begin{problem}[Tolerance Verification]
\label{problem:tolver}
Tolerance verification is defined as a predicate function with three arguments; two objects,  $\Psi_\mathcal{I}$ and $\mathcal{B}$, and a tolerance, $\Delta > 0$. The function evaluates the inequality
\begin{equation}\label{eq:tolver}
    d_\mathrm{min} (\Psi_\mathcal{I}, \mathcal{B}) > \Delta
\end{equation}
 and returns either $true$ or $false$.
\cref{fig:defn-tolver} depicts a scenario, when a curve is separated from a compact set by a distance greater than $\Delta$.
\end{problem}


\begin{problem}[Collision Detection]
\label{problem:coldet}
The collision detection function is defined as a predicate function with two arguments. These arguments are two objects,  $\Psi_\mathcal{I}$ and $\mathcal{B}$. The function determines the truth of the following statement:
\begin{equation}\label{eq:coldet}
\Psi_\mathcal{I} \cap \mathcal{B} \neq \emptyset.
\end{equation}
\end{problem}

\begin{remark}
Notice that the solution of \cref{problem:mindist} implies the solution of \cref{problem:tolver}, which in turn implies the solution of \cref{problem:coldet}. Nevertheless, defining each problem by itself is useful because, as we will see in \cref{sec:alg}, the numerical methods can be specifically tailored for each problem in order to improve its computational efficiency.
\end{remark}

\section{Bounding Methods and Analysis}
\label{sec:bounds}
The problems discussed in \cref{sec:prob} are generally non-convex and difficult to solve\footnote{Non-convex problems are considered to be at least NP-hard.}. Additionally, when a feasible solution is found it is still difficult to verify that the solution is indeed the global minimum. However, methods that rely on branch-and-bound and interval analysis \cite{jaulin2001} obtain a solution with a certificate of its optimality within some prescribed accuracy. This is achieved by successively \textit{branching} the problem into smaller and smaller sub-problems over which \textit{bounds} on the optimal solution are computed through relaxation. In this section we present the results and the analysis behind the relaxation of \cref{problem:mindist} and its sub-problems.

\subsection{Upper Bound on the Length of a Curve}
As the presented methods require the computation of the arc length of parametric curves, we will only consider parametric curves that are rectifiable \ie \ they possess a finite arc length over their domain. The absolute continuity of $\psi$ over $\mathcal{I}$ is a sufficient condition \cite{iseki1960} for the curve to be rectifiable\footnote{More generally, any function of bounded variation is rectifiable \cite{kolmogorov1957}, however, the derivatives for such functions may not exist almost everywhere.}, and the corresponding arc length function over the interval $\mathcal{Q} \subseteq \mathcal{I}$ is defined as
\begin{equation}
s_\psi(\mathcal{Q}) = \int_{\mathcal{Q}} \|\psi^\prime(t)\| \ \dt, \label{eq:arclength}
\end{equation}
where $\psi$ has a finite derivative $\psi^\prime$ almost everywhere and is Lebesgue integrable.

As branch-and-bound techniques involve repeated calculations on the curve, for reasons having to do with computational efficiency, it is extremely desirable to have a closed-form expression for the antiderivative of the square root of the inner product of $\psi'$. However, the arc length of rectifiable parametric curves cannot be obtained in general. Even for simple curves described with polynomial or sinusoidal basis functions, one cannot express the integral in terms of elementary functions. In the following result, we introduce an upper bound on $s_\psi(\mathcal{Q})$ in \cref{eq:arclength} that is better suited to provide a closed-form expression.

\begin{lemma}\label{lemma:arclength}
Let $\psi$ be an absolutely continuous function and have a derivative $\psi^\prime$ on the compact interval $\mathcal{I}$. For any closed $\mathcal{Q} \subseteq \mathcal{I}$ define the upper bound
\begin{equation}\label{eq:arclength_ub}
u_\psi(\mathcal{Q}) = \sqrt{|\mathcal{Q}|\int_\mathcal{Q} \psi^\prime(t)^\top \psi^\prime(t) \dt},
\end{equation}
where $|\mathcal{Q}|$ is the length of the sub-interval. Then for the arc length function of the curve $s_\psi$ (from \cref{eq:arclength}) the following inequality holds
\[
s_\psi(\mathcal{Q}) \le u_\psi(\mathcal{Q}).
\]
\end{lemma}
\begin{proof}
Consider the arc length function given in ~\cref{eq:arclength}. Expanding the Euclidean norm and scaling both sides yields
\[
\frac{1}{|\mathcal{Q}|} s_\psi(\mathcal{Q}) = \frac{1}{|\mathcal{Q}|} \int_\mathcal{Q} \sqrt{\psi^\prime(t)^\top \psi^\prime(t)} \dt.
\]
Note that $\psi^\prime(t)^\top \psi^\prime(t)$ is a strictly positive real-valued function and that the square root function is concave on the interval $[0, \infty)$. Using Jensen's inequality we have
\[
\frac{1}{|\mathcal{Q}|} \int_\mathcal{Q} \sqrt{\psi^\prime(t)^\top \psi^\prime(t)} \dt \le
\sqrt{\frac{1}{|\mathcal{Q}|}\int_\mathcal{Q} \psi^\prime(t)^\top \psi^\prime(t) \dt}.
\]
Multiplying by $|\mathcal{Q}|$ on both sides gives  the result.
\end{proof}

For common parametric curves with polynomial or trigonometric basis functions, such as the ones typically employed in trajectory generation methods \cite{van2015}, the antiderivative for the inner product of $\psi'$ is readily available, which greatly reduces the computation time. In \cref{tab:arclength}, notice the improvement in computation time when the antiderivative is known in closed-form.

\begin{table}[htbp]
    \small
    \centering
    \begin{tabular}{  c||cc}
       \toprule \toprule
        \multicolumn{1}{c||}{\multirow{2}{*}{\textbf{Parametric Equations}}} & \multicolumn{2}{c}{\textbf{Median Time} (ns)}  \\
        \\[-1em]
        \cline{2-3}\\[-1em]
        & $s_\psi(\mathcal{Q})$ & $u_\psi(\mathcal{Q})$ \\
        \hline \hline \\[-1em]
        $(2\cos(t), \enskip \sin(t))$ & 1710.70 & 58.02\\
        \hline \\[-1em]
        $((t^3 + t), \enskip t)$ & 1673.80 & 44.69\\
        \hline \\[-1em]
        $((t + 1)^{-1}, \enskip t)$ & 2102.11 & 38.43\\
        \toprule
    \end{tabular}
    \caption{Comparison between the median evaluation time for computing $s_\psi(\mathcal{Q})$ and $u_\psi(\mathcal{Q})$ for several different intervals $\mathcal{Q} \subseteq [0, 1]$. For each example presented in the table, the antiderivative of the integrand in \cref{eq:arclength} is not available in closed form and must be numerically integrated, whereas the antiderivative of the integrand in \cref{eq:arclength_ub} is available in closed-form. The numerical integration uses the Gauss-Kronrod quadrature formula procedure over 15 points.}
    \label{tab:arclength}
\end{table}

\subsection{Convex Hull on Sub-intervals of a Curve}
The following theorem establishes a convex hull for parametric curves on any sub-interval based on their upper bound of the arc length from \cref{eq:arclength_ub}.

\begin{theorem}[Convex Hull] \label{theorem:convex}
Let $\psi$ be an absolutely continuous function  defined on the compact interval $\mathcal{I}$. For any closed interval $[\alpha, \beta] \equiv \mathcal{Q} \subseteq \mathcal{I}$, define the convex compact set
\begin{equation} \label{eq:bound}
    \mathcal{U}_{\mathcal{Q}}= \{ x \in \mathbb{R}^d :
    \|\psi(\alpha) - x \| + \|x - \psi(\beta)\| \le u_{\psi}(\mathcal{Q})
    \}.
\end{equation}
Then the curve defined on the interval $\mathcal{Q}$ satisfies
\begin{equation} \label{eq:ellipsoidx}
    \Psi_\mathcal{Q} \subset \mathcal{U}_\mathcal{Q}.
\end{equation}
\end{theorem}
\begin{proof}
It is obvious that the shortest distance between two points in Euclidean space is the length of the chord joining them. Therefore  for all $t \in [\alpha, \beta]$ the following must hold
\[
\|\psi(\alpha) - \psi(t)\| + \|\psi(t) - \psi(\beta)\| \le s_{\psi}([\alpha, t]) + s_{\psi}([t, \beta]).
\]
Then, from the definition of the arc length  in \cref{eq:arclength} we have that $s_\psi([\alpha, t]) + s_{\psi}([t, \beta])$ is the sum of integrals on adjacent intervals. Since $\psi$ is rectifiable, the integrals are combined as the total arc length on $\mathcal{Q}$, implying that
\[
\|\psi(\alpha) - \psi(t)\| + \|\psi(t) - \psi(\beta)\| \le s_{\psi}(\mathcal{Q}).
\]
Recall that since $s_\psi(\mathcal{Q}) \le u_\psi(\mathcal{Q})$ from \cref{lemma:arclength}, then every point on the curve evaluated over the interval $\mathcal{Q}$ is an element of the set $\mathcal{U}_\mathcal{Q}$, \ie
\[
\psi(t) \in \mathcal{U}_\mathcal{Q}.
\]
From the definition of the curve in \cref{eq:curve} it follows that $\Psi_\mathcal{Q} \subset  \mathcal{U}_\mathcal{Q}$.
\end{proof}

\begin{figure}[ht]
  \centering
  \includegraphics[width=0.75\columnwidth]{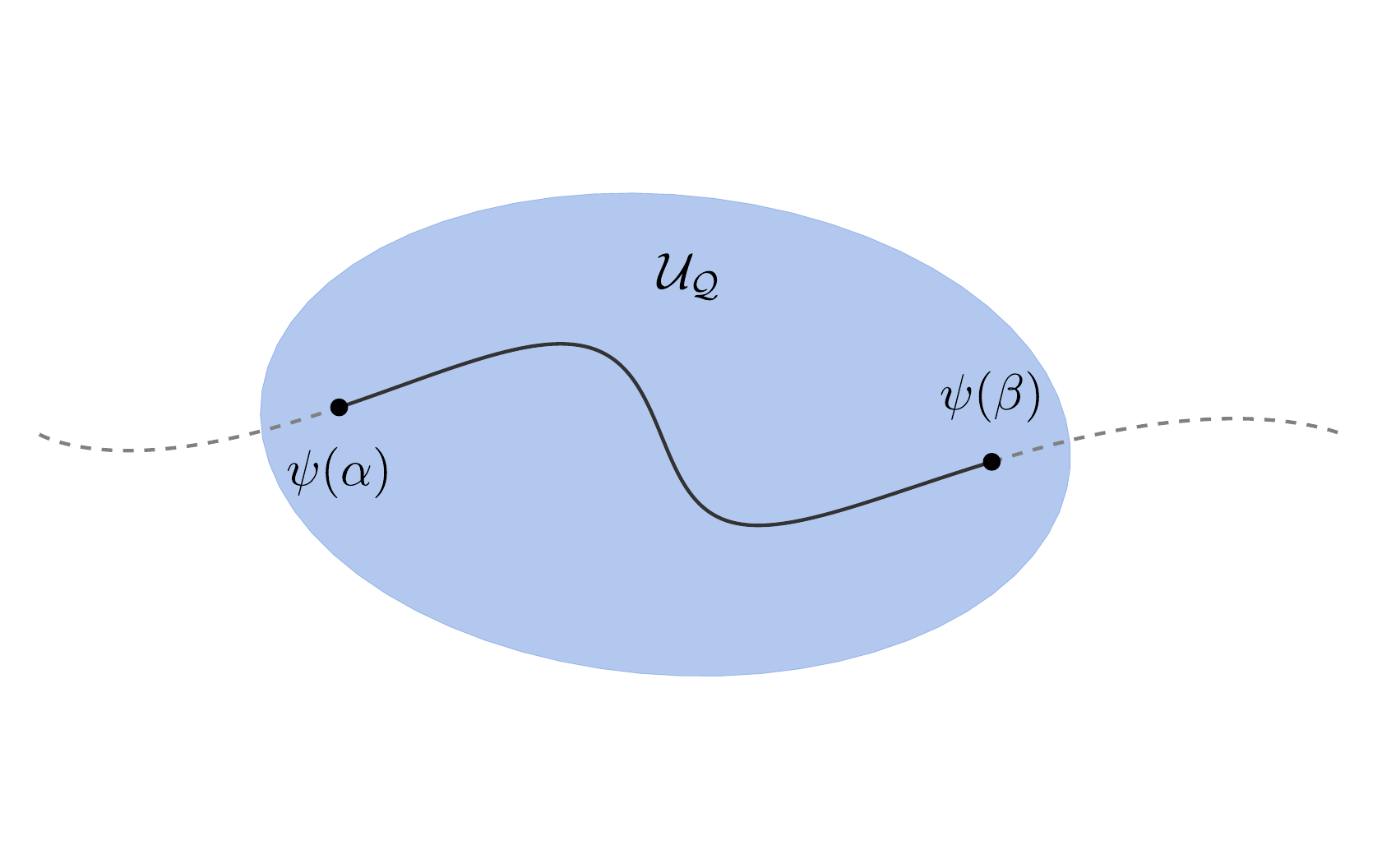}
  \caption{The blue shaded region shows the convex set $\mathcal{U}_{\mathcal{Q}}$, which contains the function $\psi$ evaluated over the sub-interval $[\alpha, \beta]$. The dashed-line is the evaluation of $\psi$ over its entire domain.}
  \label{fig:ellipse}
\end{figure}

The convex hull in \cref{theorem:convex} is a compact interval for curves in $\mathbb{R}$, and is also an ellipsoid for curves in higher dimensions. This geometric relationship to an ellipsoid is particularly useful because there are several methods \cite{gilbert1988, rimon1997, chakraborty2008} to cheaply compute the minimum separating distance to an ellipsoid. For a convex hull $\mathcal{U}_{\mathcal{Q}}$ of the function $\psi$ evaluated over a sub-interval $[\alpha, \beta] \equiv \mathcal{Q} \subseteq \mathcal{I}$, the foci of the ellipse are at $\psi(\alpha)$ and $\psi(\beta)$. The major axis has a length of $u_\psi(\mathcal{Q})$, and the minor axes are of equal lengths $\sqrt{u_\psi(\mathcal{Q})^2 - \|\psi(\alpha) -  \psi(\beta)\|^2}$. This construction is illustrated in \cref{fig:ellipse}.

\begin{remark}
Fat arcs \cite{sederberg1989, bartovn2011} provide tighter bounds with cubic convergence, however, the non-convexity of the bounding region is problematic when computing proximity queries. Additionally, they are only useful in the context of planar \Bez curves and spirals. The convergence behavior of the ellipsoidal bounding region to the curve will be explored in the future.
\end{remark}

\subsection{Relaxation of \cref{problem:mindist}}
\begin{figure}[ht]
  \centering
  \subfloat[]{\label{fig:ellipse_lba}\includegraphics[width=0.5\columnwidth]{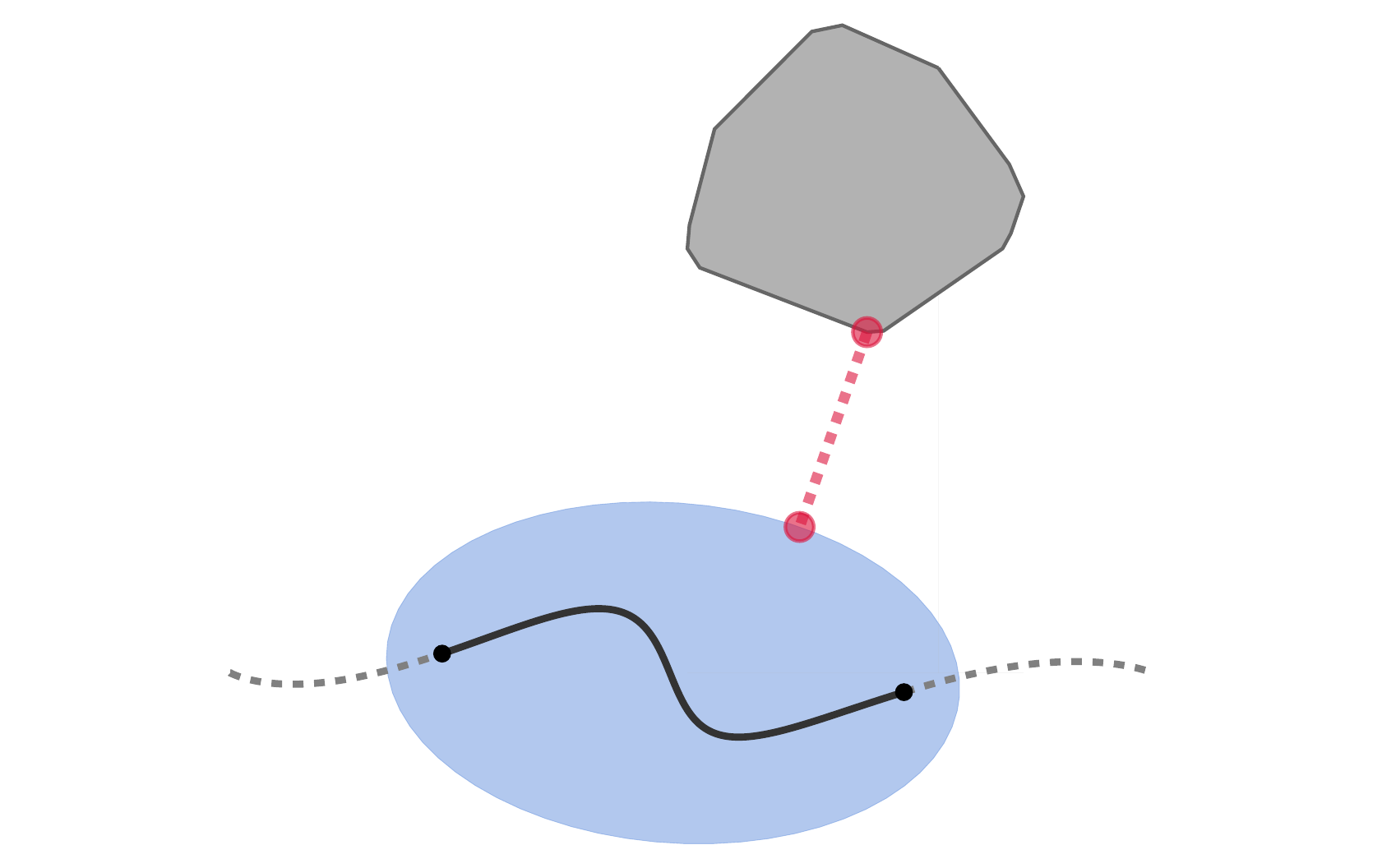}}
  \subfloat[]{\label{fig:ellipse_lbb}\includegraphics[width=0.5\columnwidth]{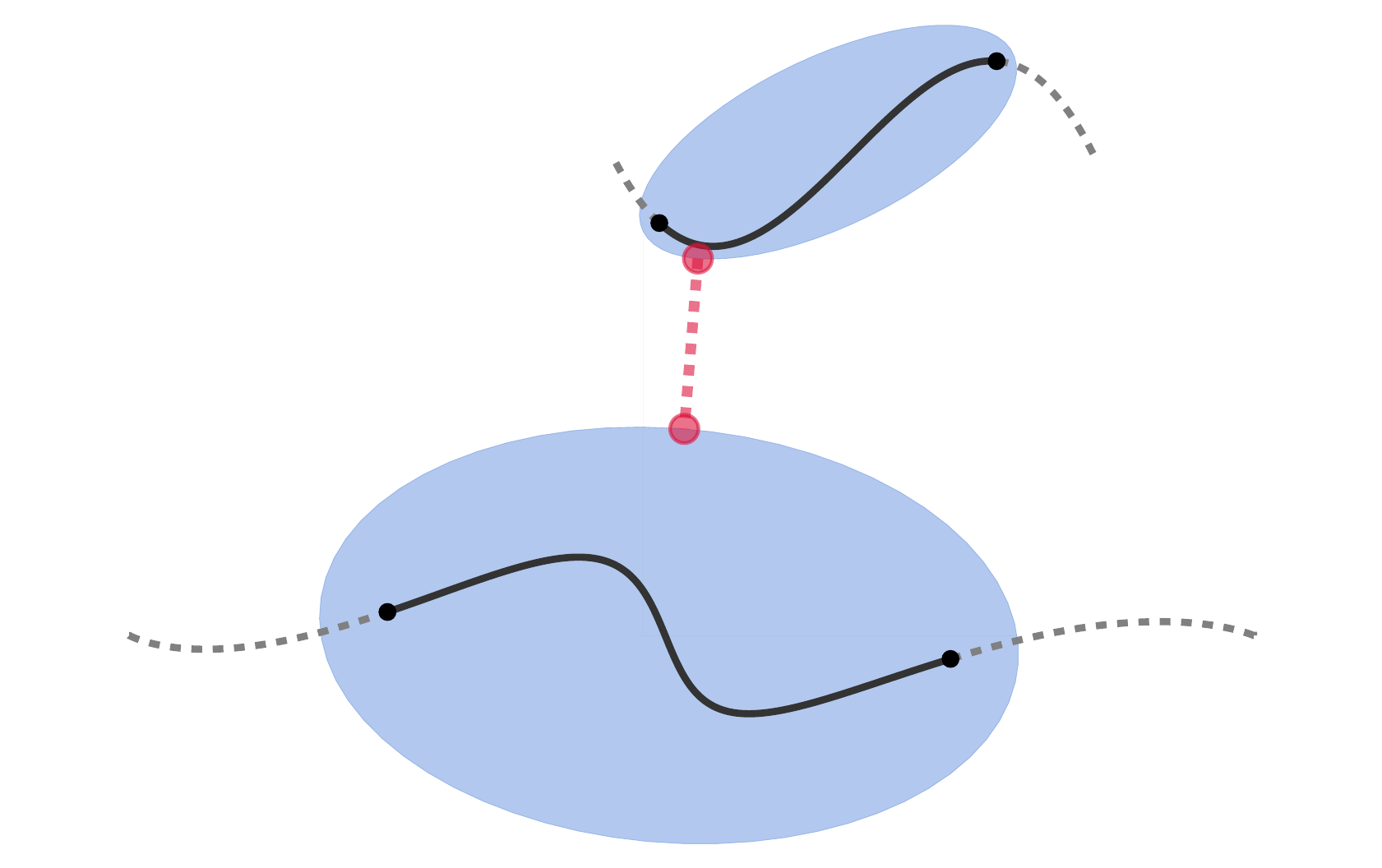}}
  \caption{Lower bound on minimum separating distance between (a) a parametric curve and a convex polygon, and (b) two parametric curves.}
\end{figure}

Now that we have enclosed a parametric curve inside a convex hull, \cref{problem:mindist} can be relaxed, and a more tractable problem can be solved by computing the minimum separating distance to the convex hull. The solution to this relaxed problem provides a lower bound to the optimal solution of the original problem. We define this lower bound on the minimum separating distance $\dlb$ between $\Psi_\mathcal{Q}$  and a compact set $B$ as
\begin{equation}\label{eq:lowerbound}
\dlb(\Psi_\mathcal{Q}, \mathcal{B}) = \min_{x \in \mathcal{U}_{\mathcal{Q}}, b \in \mathcal{B}} \| x - b \|,
\end{equation}
where $\mathcal{U}_{\mathcal{Q}}$ is the convex hull corresponding to $\Psi_\mathcal{Q}$ from \cref{eq:bound}. \cref{fig:ellipse_lba} shows the lower bound between a convex polygon and a parametric curve evaluated over a sub-interval. As one might expect, if the object $\mathcal{B}$ is also a parametric curve, then $\dlb$ is computed between the convex sets constructed from each of the curves (as shown in \cref{fig:ellipse_lbb}), as
\begin{equation}\label{eq:lowerbound_curve}
\dlb(\Psi_\mathcal{Q}, \Phi_\mathcal{R}) = \min_{x \in \mathcal{U}_{\mathcal{Q}}, y \in \mathcal{V}_{\mathcal{R}}} \| x - y \|,
\end{equation}
where $\Phi_\mathcal{R}$ is the parametric curve $\phi: \mathbb{R} \supset \mathcal{R} \to \mathbb{R}^d$, and $\mathcal{V}_\mathcal{R}$ is the convex hull of $\Phi_\mathcal{R}$.

\begin{figure}[ht]
  \centering
  \subfloat[]{\includegraphics[width=0.5\columnwidth]{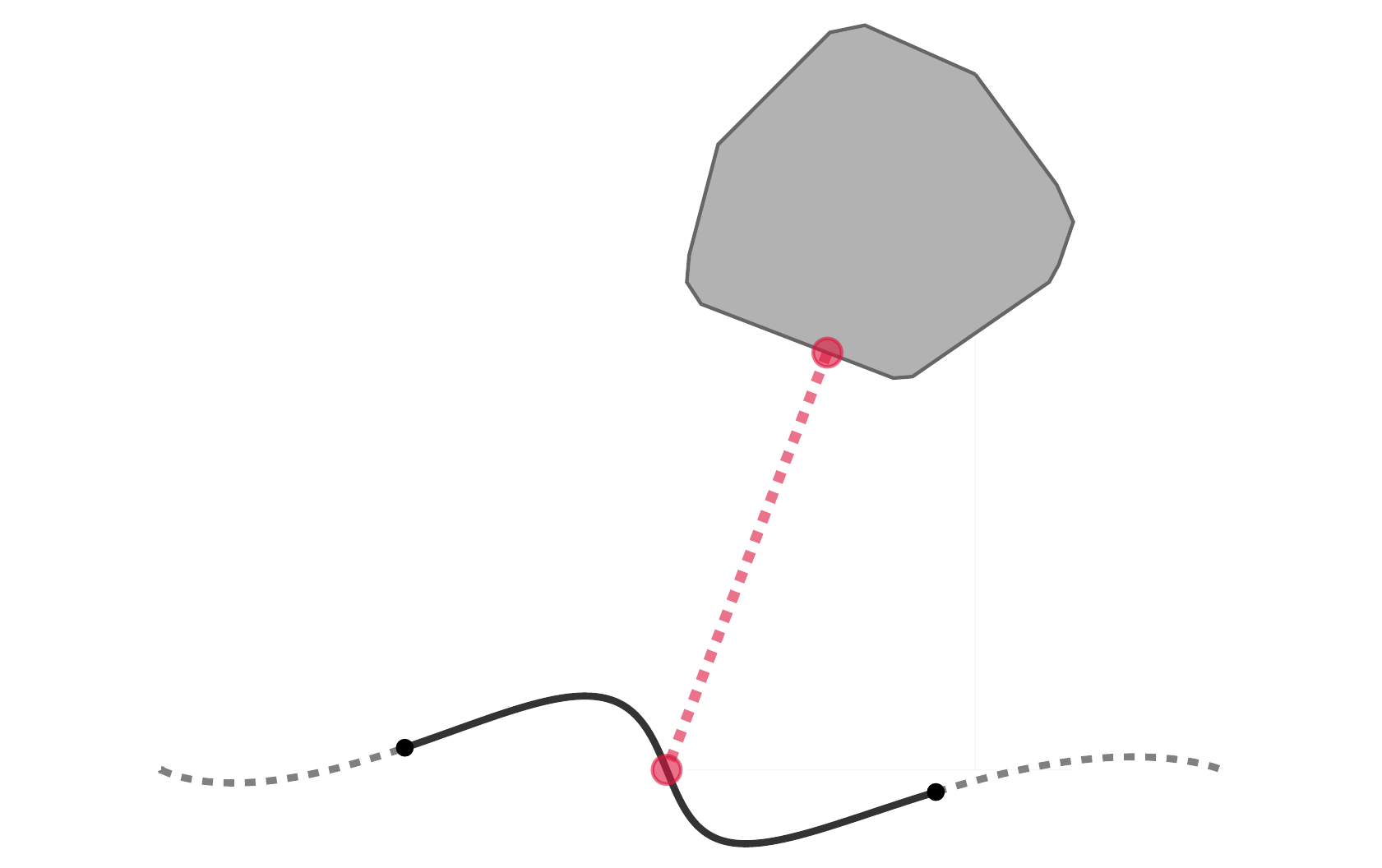}}
  \subfloat[]{\includegraphics[width=0.5\columnwidth]{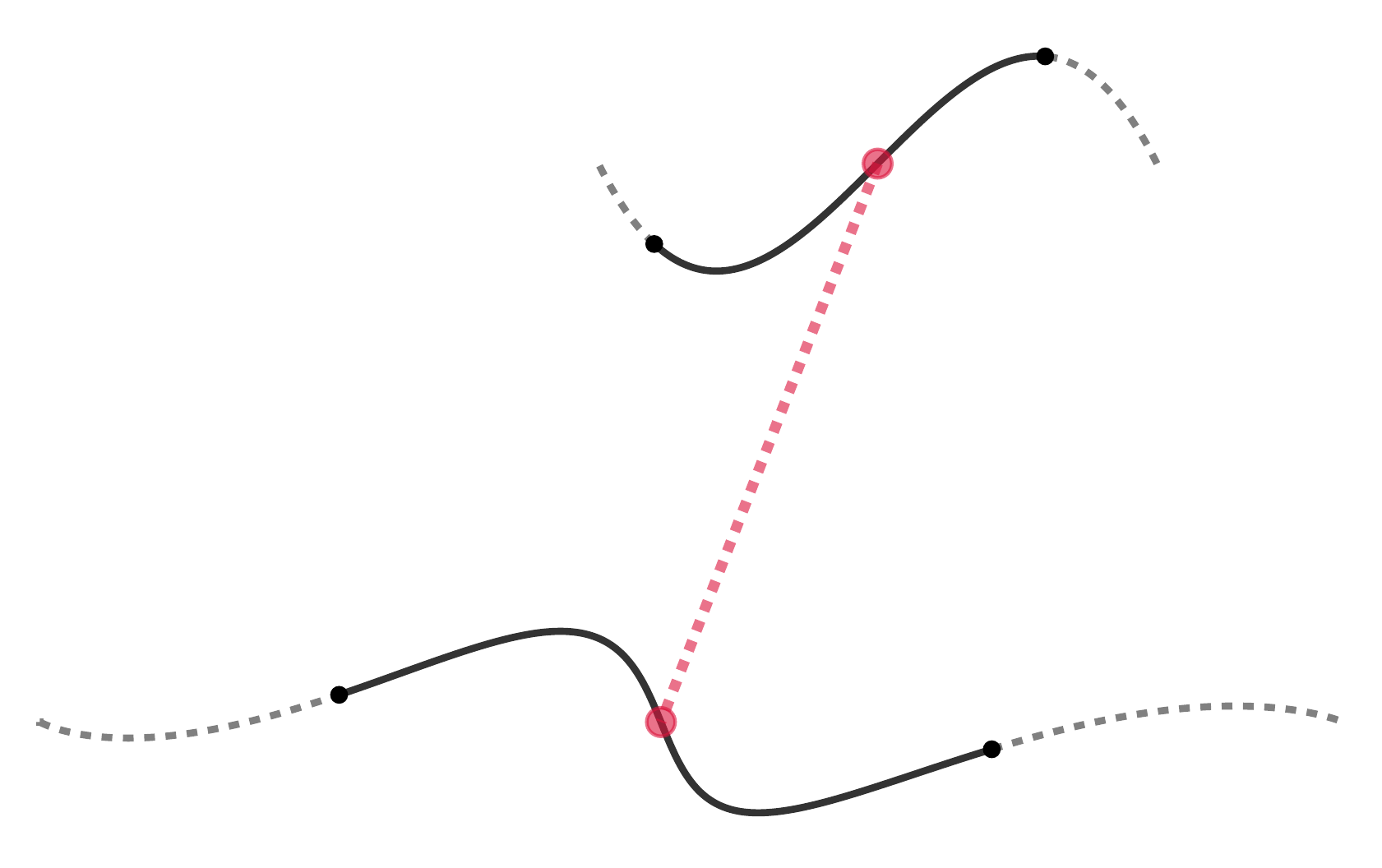}}
  \caption{Upper bound on minimum separating distance between (a) a parametric curve and a convex polygon, and (b) two parametric curves.}
  \label{fig:ellipse_ub}
\end{figure}

In addition to the lower bound, an upper bound on the optimal solution must also be defined in order to bound the solution in a compact interval. The upper bound on the minimum separating distance $\dub$ between $\Psi_\mathcal{Q}$  and a compact set $B$ is given by
\begin{equation}\label{eq:upperbound}
\dub(\Psi_\mathcal{Q}, B) = \min_{b \in \mathcal{B}} \| x'(\Psi_\mathcal{Q}) - b \|,
\end{equation}
 such that $x'(\Psi_\mathcal{Q}) \in \Psi_\mathcal{Q}$. Similarly, $\dub$ between the two parametric curves $\Psi_\mathcal{Q}$ and $\Phi_\mathcal{R}$ is
\begin{equation}\label{eq:upperbound_curve}
\dub(\Psi_\mathcal{Q}, \Phi_\mathcal{R}) = \| x'(\Psi_\mathcal{Q}) - x''(\Phi_\mathcal{R}) \|,
\end{equation}
such that $x'(\Psi_\mathcal{Q}) \in \Psi_\mathcal{Q}$ and $x''(\Phi_\mathcal{R}) \in \Phi_\mathcal{R}$. In the numerical implementation as seen in \cref{fig:ellipse_ub}, the functions $x'$ and $x''$ select the middle points of $\Psi_\mathcal{Q}$ and $\Phi_\mathcal{R}$ respectively. Heuristics to select points in the curve set based on the relative configuration of the objects rather than simply the midpoints may find tighter bounds, however, this is beyond the scope of the paper.

We present analysis on the functions $\dlb$ and $\dub$. We show for each interval $\mathcal{Q} \subseteq \mathcal{I}$ that indeed these functions bound the optimal solution on that interval, and that the difference in upper and lower bounds converges uniformly to zero as $|Q| \rightarrow 0$. The boundedness and uniform convergence results are necessary in order to guarantee the convergence of interval branch-and-bound algorithms in finite time.

\begin{theorem}[Boundedness]\label{theorem:bound}
Let $\mathcal{B}$ be a compact set in $\mathbb{R}^d$, and $\Psi_\mathcal{I}$ be the curve $\psi: \mathcal{I} \to \mathbb{R}^d$. Then, for every closed $\mathcal{Q} \subseteq \mathcal{I}$, we have
\begin{equation}
\dlb(\Psi_\mathcal{Q}, \mathcal{B}) \le
\dmin(\Psi_\mathcal{Q}, \mathcal{B}) \le
\dub(\Psi_\mathcal{Q}, \mathcal{B}).
\end{equation}
\end{theorem}
\begin{proof}
It is clear that for all $t' \in \mathcal{Q} \subseteq \mathcal{I}$
\begin{equation*}
    \min_{t\in\mathcal{Q}, b \in \mathcal{B}} \|\psi(t) - b\| \le
    \min_{b \in \mathcal{B}} \|\psi(t') - b\|,
\end{equation*}
which proves the result that $\dmin(\Psi_\mathcal{Q}, \mathcal{B}) \le \dub(\Psi_\mathcal{Q}, \mathcal{B})$. Recall from \cref{theorem:convex} that $\Psi_\mathcal{Q} \subset \mathcal{U}_\mathcal{Q}$, which implies
\begin{equation*}
    \min_{a\in\mathcal{U}_\mathcal{Q}, b \in \mathcal{B}} \|a - b\| \le
    \min_{a\in\Psi_\mathcal{Q}, b \in \mathcal{B}} \|a - b\|.
\end{equation*}
Thus, $\dlb(\Psi_\mathcal{Q}, \mathcal{B}) \le \dmin(\Psi_\mathcal{Q}, \mathcal{B})$, and we have the result.
\end{proof}

\begin{theorem}[Uniform Convergence] \label{theorem:convergence}
Let $\mathcal{B}$ be a compact set in $\mathbb{R}^d$, and $\Psi_\mathcal{I}$ be the curve $\psi: \mathcal{I} \to \mathbb{R}^d$. For every $\epsilon > 0$, there exists a $\delta > 0$ such that for all $\{\mathcal{Q} : \mathcal{Q} \subseteq \mathcal{I}, |\mathcal{Q}| < \delta\}$, we satisfy
\begin{equation}
\dub(\Psi_\mathcal{Q}, \mathcal{B}) -
\dlb(\Psi_\mathcal{Q}, \mathcal{B}) < \epsilon.
\end{equation}
\end{theorem}
\begin{proof}
For any $\mathcal{Q} \subseteq \mathcal{I}$ recall that
\begin{equation*}
\dub(\Psi_\mathcal{Q}, \mathcal{B}) - \dlb(\Psi_\mathcal{Q}, \mathcal{B}) = 
\min_{b \in \mathcal{B}}\|x'(\Psi_\mathcal{Q}) - b\| - \|x^* - b^*\|,
\end{equation*}
such that $x^* \in \mathcal{U}_\mathcal{Q}$ and $b^* \in \mathcal{B}$  minimize \cref{eq:lowerbound}. Since $\min_{b \in\mathcal{B}} \| x'(\Psi_\mathcal{Q}) - b \|$ will be bounded from above by any other element in the set $\mathcal{B}$, we have the following inequality:
\[
\dub(\Psi_\mathcal{Q}, \mathcal{B}) - \dlb(\Psi_\mathcal{Q}, \mathcal{B}) \le \|x'(\Psi_\mathcal{Q}) - b^*\| - \|x^* - b^*\|.
\]
From the reverse triangle inequality we have
\[
\dub(\Psi_\mathcal{Q}, \mathcal{B}) - \dlb(\Psi_\mathcal{Q}, \mathcal{B}) \le \|x'(\Psi_\mathcal{Q}) - x^*\|.
\]
Then from \cref{theorem:convex}, we know that the largest variation of elements in $\mathcal{U}_\mathcal{Q}$ is bounded by the upper bound of the arc length, $u_\psi(\mathcal{Q})$. And, since $x'(\Psi_\mathcal{Q}) \in \Psi_\mathcal{Q} \subset \mathcal{U}_\mathcal{Q}$ and $x^* \in \mathcal{U}_\mathcal{Q}$, we get
\[
\dub(\Psi_\mathcal{Q}, \mathcal{B}) - \dlb(\Psi_\mathcal{Q}, \mathcal{B}) \le u_\psi(\mathcal{Q}). 
\]
From \cref{eq:arclength_ub}, we expand $u_\psi(\mathcal{Q})$ with the expression
\begin{equation*}
\dub(\Psi_\mathcal{Q}, \mathcal{B}) - \dlb(\Psi_\mathcal{Q}, \mathcal{B}) \le  \sqrt{|\mathcal{Q}|\int_\mathcal{Q}\psi'(t)^\top \psi'(t) \dt}.
\end{equation*}
Since $\psi$ is real valued, the inner product of $\psi'$ is strictly positive. Thus, we further upper bound our expression by expanding the limits of integration with $\mathcal{I} \supseteq \mathcal{Q}$. Define $0 < k = \int_\mathcal{I}\psi'(t)^\top \psi'(t) \dt$. Then we have
\begin{equation*}
\dub(\Psi_\mathcal{Q}, \mathcal{B}) - \dlb(\Psi_\mathcal{Q}, \mathcal{B}) \le  \sqrt{|\mathcal{Q}|k}.
\end{equation*}
Thus, for every $\epsilon > 0$, choose $\delta = \frac{\epsilon^2}{k}$ so that for all $\mathcal{Q} \subseteq \mathcal{I}$ such that $|\mathcal{Q}| < \delta$ we get
\begin{equation*}
\dub(\Psi_\mathcal{Q}, \mathcal{B}) - \dlb(\Psi_\mathcal{Q}, \mathcal{B}) < \epsilon.
\end{equation*}
\end{proof}

\begin{remark}
The minimization problems of \cref{eq:lowerbound,eq:upperbound} may also be nonconvex depending on the geometry of $\mathcal{B}$. We refer readers to \cite{gilbert1988,lin1991} for convex polytopes and \cite{gottschalk1996, larsen1999} for more complex models.
\end{remark}

\section{Proximity Queries}
\label{sec:alg}
Given the boundedness results of \cref{theorem:bound} and the convergence results of \cref{theorem:convergence}, we present algorithms that evaluate the solutions to \cref{problem:mindist,problem:tolver,problem:coldet} within some tolerance. Central to each of these algorithms is a process of interval subdivision and bounding. First, we present the methods to compute the minimum separating distance between a parametric curve and a compact set (or another parametric curve). Later, we highlight the variations of this algorithm for evaluating the tolerance verification and collision detection queries without computing the minimum separating distance between the objects.

\subsection{Minimum Separating Distance}
Given a parametric function $\psi: \mathcal{I} \to \mathbb{R}^d$, \cref{alg:mindist} computes the minimum separating distance between the curve $\Psi_\mathcal{I}$ and a compact set $\mathcal{B}$.

\begin{algorithm}[htbp]
    \caption{Minimum Separating Distance $(\Psi_\mathcal{I}, B)$} \label{alg:mindist}
    \begin{algorithmic}[1]
        \State $\mathcal{L} \gets \{ \mathcal{I} \}$\label{alg:leaves_setup}
        \State $\overline{d} \gets \dub(\Psi_\mathcal{I}, \mathcal{B})$\label{alg:ubsetup}
        \State $\underline{d}  \gets \dlb(\Psi_\mathcal{I}, \mathcal{B})$\label{alg:lbsetup3}
        \While{$\overline{d} - \underline{d} > \epsilon$}
            \State $\mathcal{X} \gets \argmin_{\mathcal{Q} \in \mathcal{L}} \dlb(\Psi_\mathcal{Q}, \mathcal{B})$
            \State $\mathcal{X}_L, \mathcal{X}_R \gets \texttt{split}(\mathcal{X})$\label{alg:split}
            \State $\mathcal{L} \gets \mathcal{L} \setminus \{\mathcal{X}\}$
            \State $\mathcal{L} \gets \mathcal{L} \cup \{\mathcal{X}_L, \mathcal{X}_R\}$
            \State $\overline{d} \gets \min_{\mathcal{Q} \in \mathcal{L}} \dub(\Psi_\mathcal{Q}, \mathcal{B})$
            \State $\underline{d}  \gets \min_{\mathcal{Q} \in \mathcal{L}} \dlb(\Psi_\mathcal{Q}, \mathcal{B})$
        \EndWhile
        \State \textbf{return} $\underline{d}$
    \end{algorithmic}
\end{algorithm}

\begin{figure}[htbp]
  \centering
  \subfloat[]{\label{fig:evolution_inital}\includegraphics[width=0.33\columnwidth]{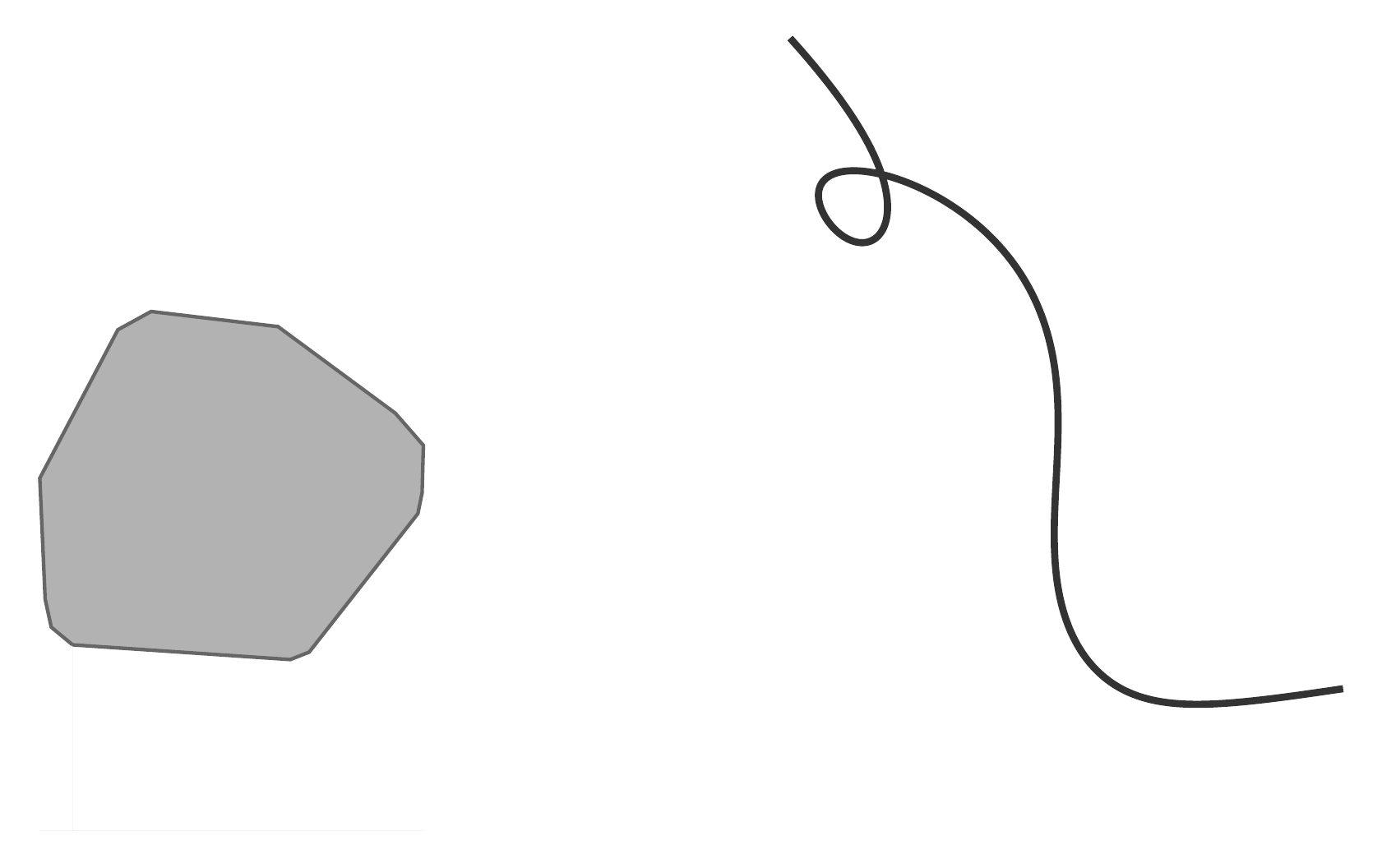}}
  \subfloat[]{\label{fig:evolution_subplots_0}\includegraphics[width=0.33\columnwidth]{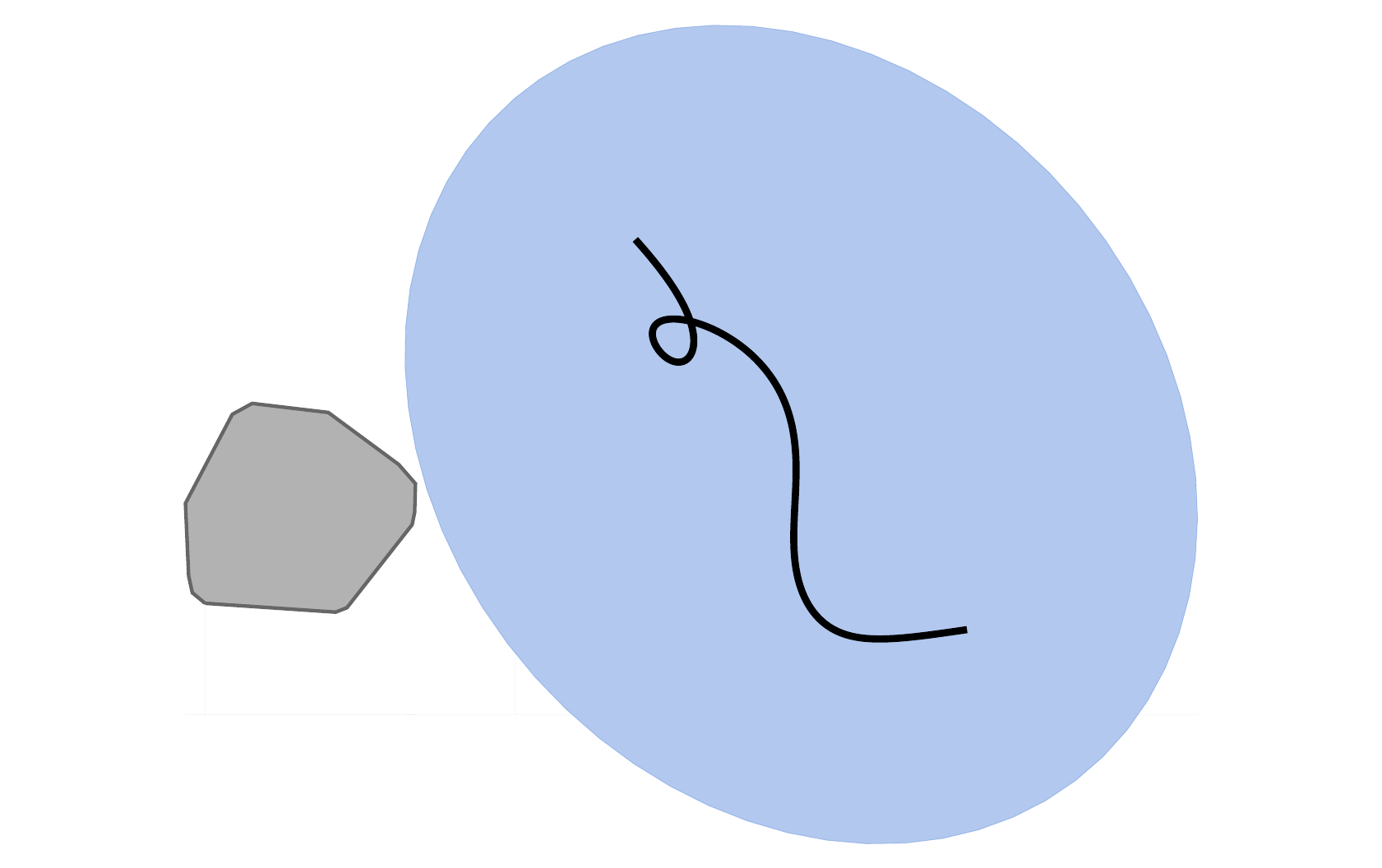}} 
  \subfloat[]{\label{fig:evolution_subplots_1}\includegraphics[width=0.33\columnwidth]{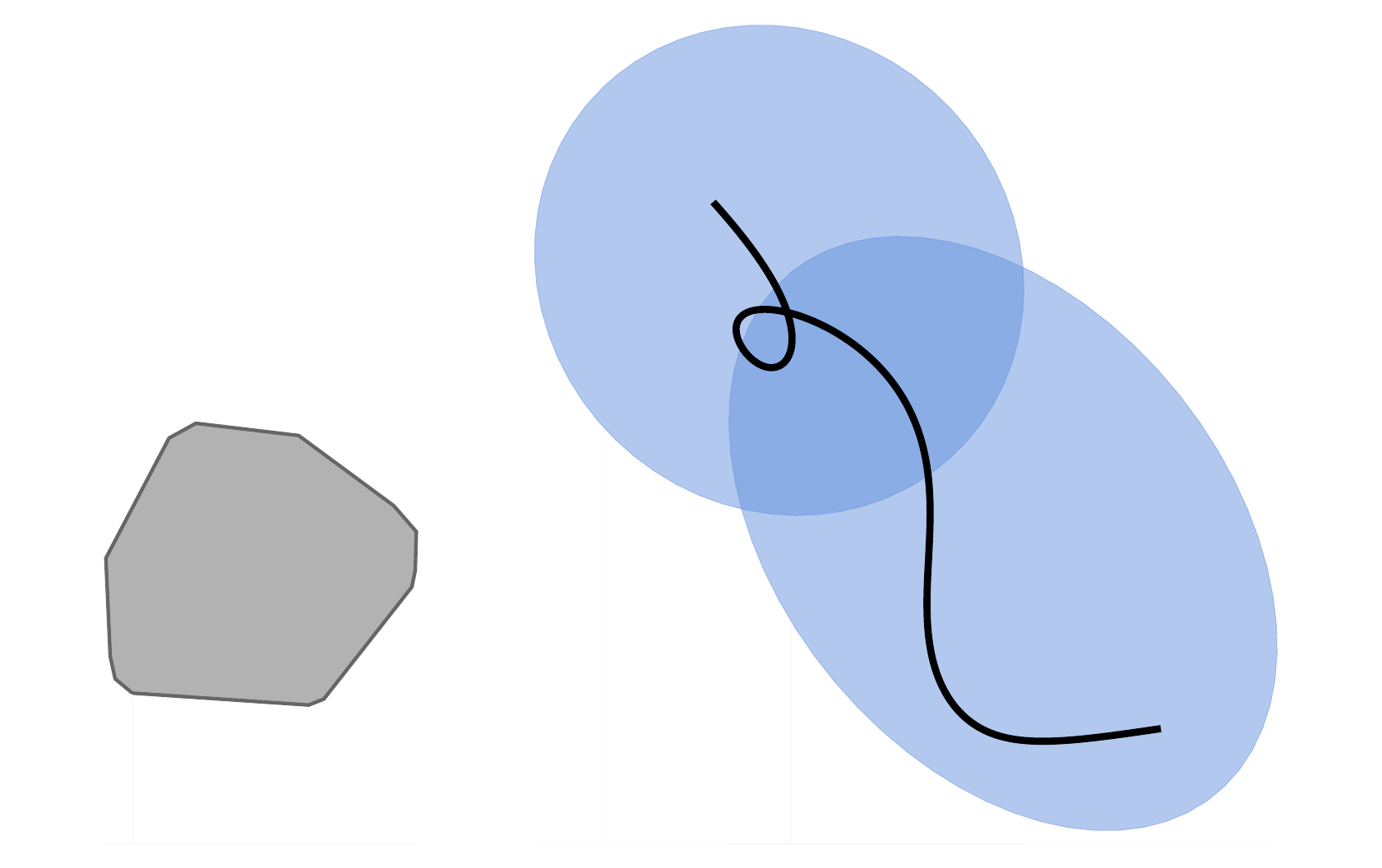}} \\
  \subfloat[]{\label{fig:evolution_subplots_3}\includegraphics[width=0.33\columnwidth]{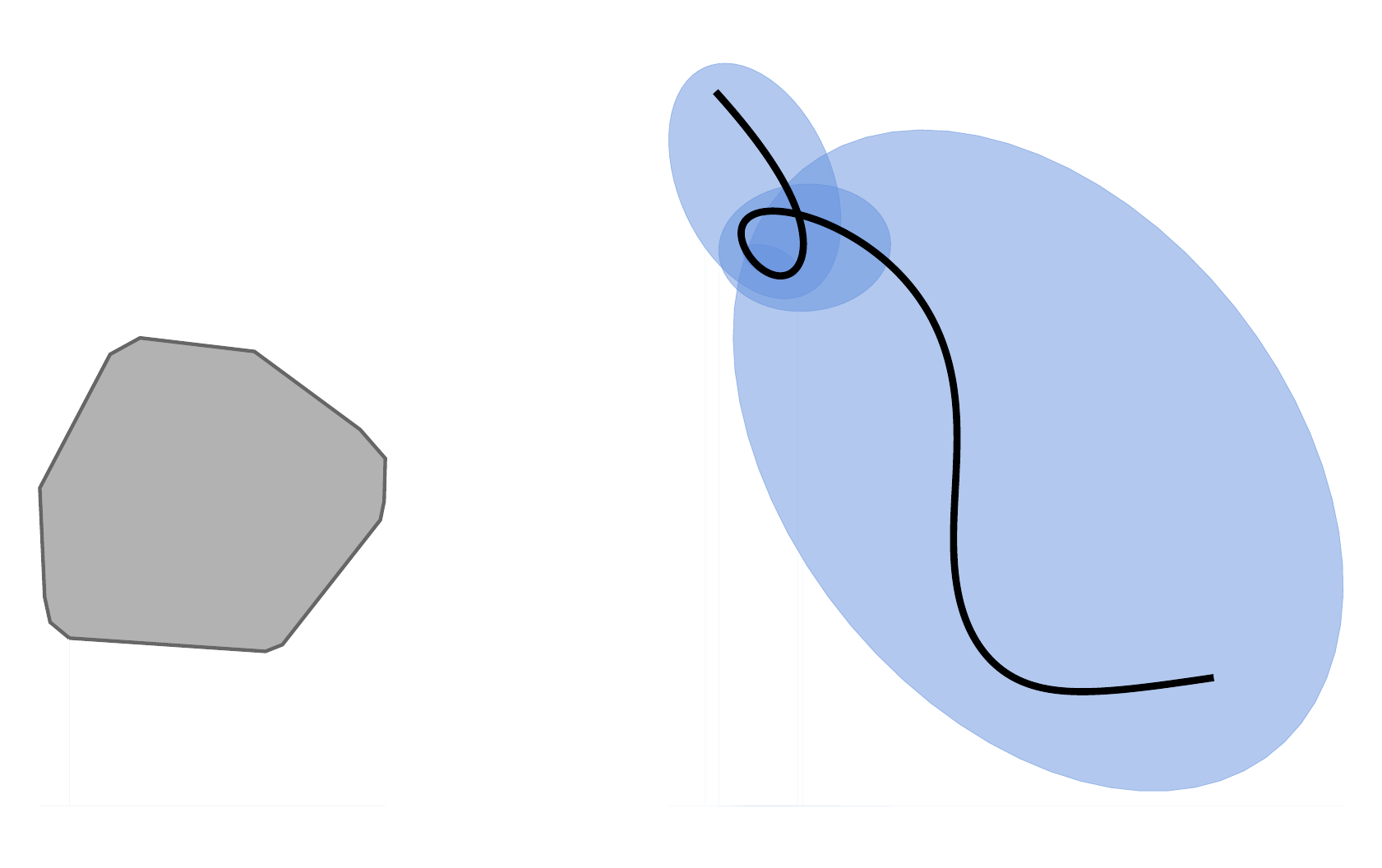}} 
  \subfloat[]{\label{fig:evolution_subplots_7}\includegraphics[width=0.33\columnwidth]{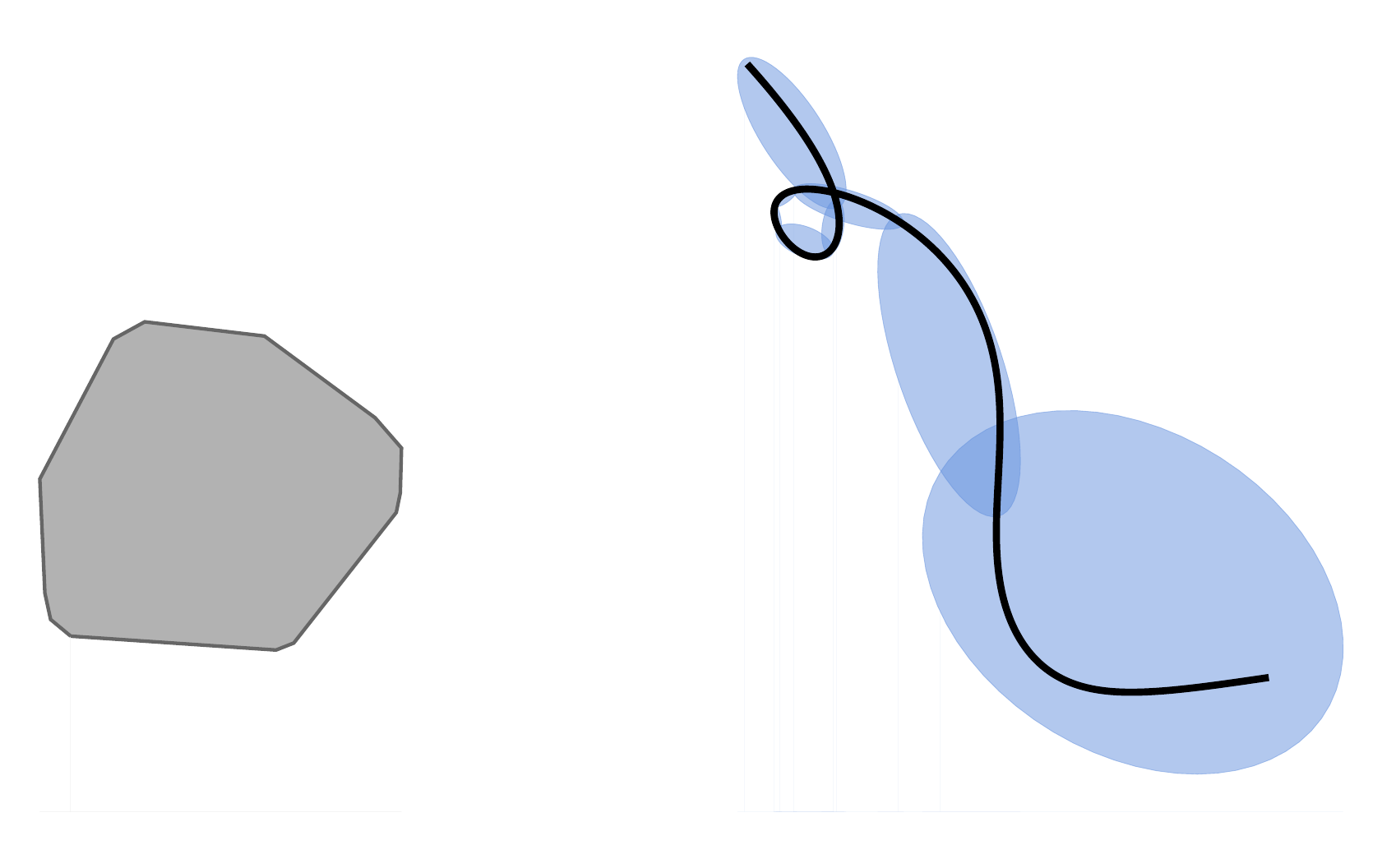}}
  \subfloat[]{\label{fig:evolution_final}\includegraphics[width=0.33\columnwidth]{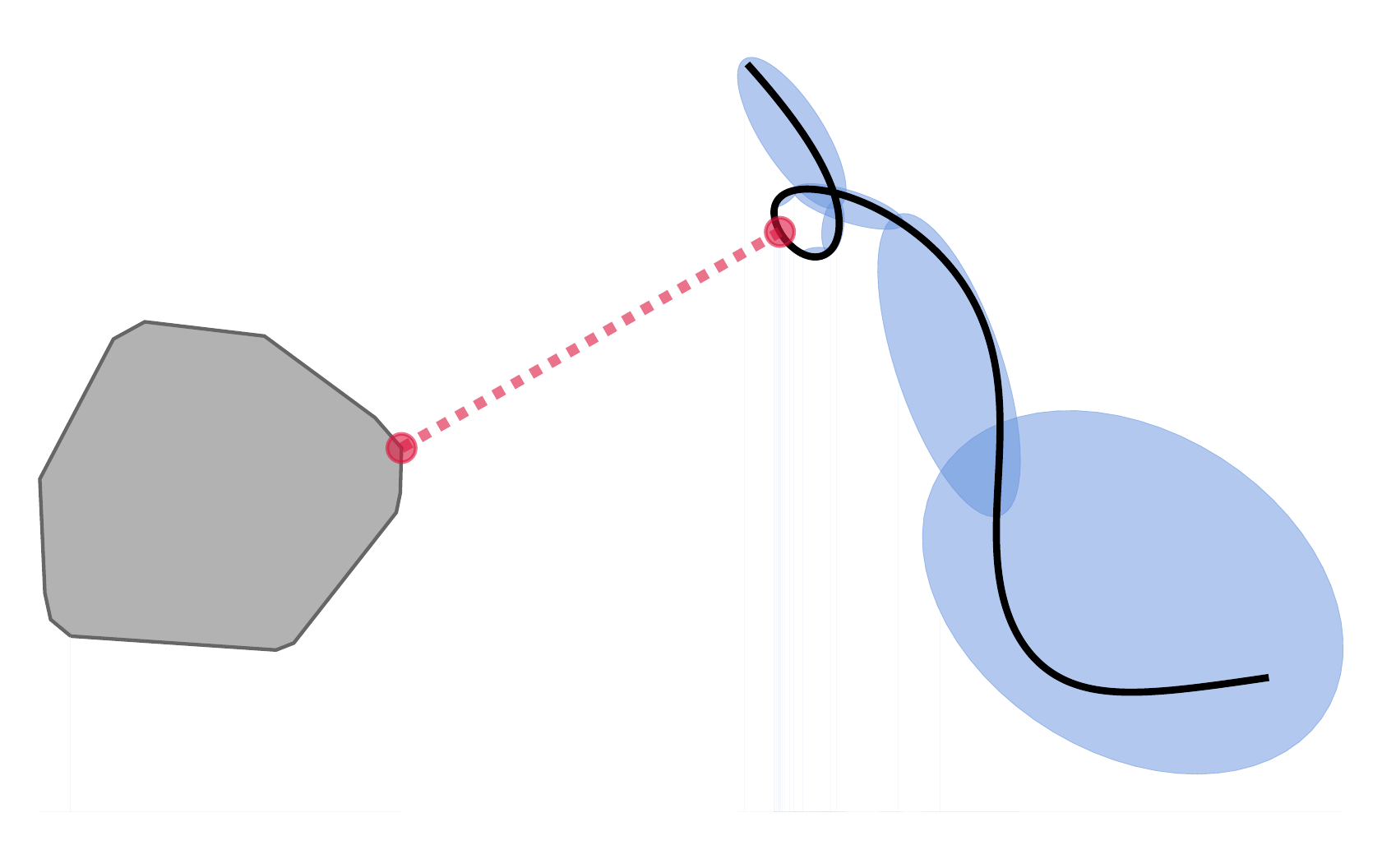}}
  \caption{The different stages of \cref{alg:mindist} when computing the minimum distance between a $\nth{13}$ order \Bez curve and a convex polygon.}
  \label{fig:evolve}
\end{figure}

We proceed to describe the algorithm. The general structure of the algorithm closely matches that of a vanilla branch-and-bound method \cite{boyd2007}. Consider computing the minimum separating distance between a curve and a convex polygon as shown in \cref{fig:evolution_inital}. A collection $\mathcal{L}$, which stores the intervals over which the bounds on the optimal solution are computed, is initialized with a single element: the entire domain $\mathcal{I}$. The relaxed problem is solved over the domain $\mathcal{I}$ (\cref{fig:evolution_subplots_0}), and the upper and lower bound values are stored in the states $\overline{d}$ and $\underline{d}$ respectively. If the difference between the bounds is above a prescribed tolerance, the algorithm proceeds to the iterative phase. $\mathcal{I}$ is split into sub-intervals $\mathcal{X}_R$ and $\mathcal{X}_L$, over which the relaxations of the problems on the new intervals are solved (\cref{fig:evolution_subplots_1}). These sub-intervals are added to the collection $\mathcal{L}$, and the original domain $\mathcal{I}$ is removed from the collection. The least upper and lower bounds on the solution are kept track of by $\overline{d}$ and $\underline{d}$, respectively, on all the sub-intervals present in the collection $\mathcal{L}$. Each successive iteration sees the sub-interval with the least lower bound chosen for subdivision. \Cref{fig:evolution_subplots_3,fig:evolution_subplots_7} show snapshots of the algorithm into the third and seventh iteration respectively. Notice that the algorithm preferentially selects sub-intervals closer to the polygon to subdivide because of the best-first search strategy.

\begin{figure}[htbp]
  \centering
  \includegraphics[width=0.75\columnwidth]{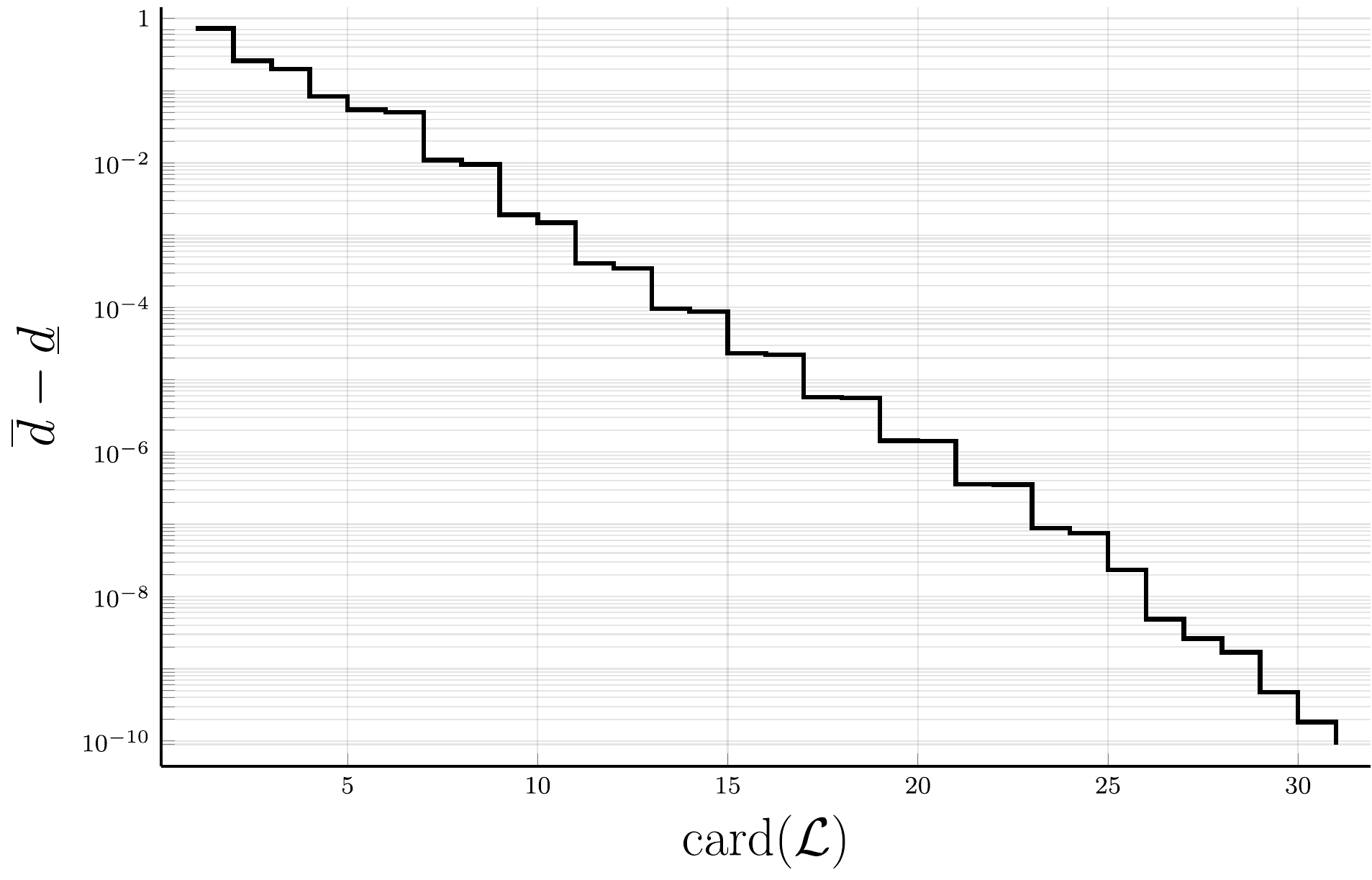}
  \caption{Convergence of $\overline{d} - \underline{d}$ for the scenario presented in \cref{fig:evolve} to $\epsilon = 10^{-10}$.}
  \label{fig:error_evolve}
\end{figure}

Since the minimization problem of \cref{eq:mindist} is over a continuous space, the branch-and-bound method will repeatedly subdivide infinitely many times (whenever the objects $\Psi_\mathcal{I}$ and $\mathcal{B}$ do not collide). However, by setting an absolute tolerance $\epsilon > 0$ on the solution, we obtain an $\epsilon$-suboptimal solution in a finite number of iterations, \ie
\begin{equation}\label{eq:interval}
\dmin(\Psi_\mathcal{I}, \mathcal{B}) \in [\underline{d}, \underline{d} + \epsilon],
\end{equation}
where $\underline{d}$ is the lower bound on the solution obtained from the minimum separating distance algorithm. The certificate proving that the global minimum lies in the interval shown in \cref{eq:interval} can be obtained by examining the collection $\mathcal{L}$. \Cref{fig:evolution_final} highlights the solution, when the bounds are separated by a magnitude of no more than $\epsilon$. The convergence of $\overline{d}-\underline{d}$ to $\epsilon$ for the example presented in \cref{fig:evolve} is shown in \cref{fig:error_evolve}.

\begin{figure}[htbp]
  \centering
  \includegraphics[width=\columnwidth]{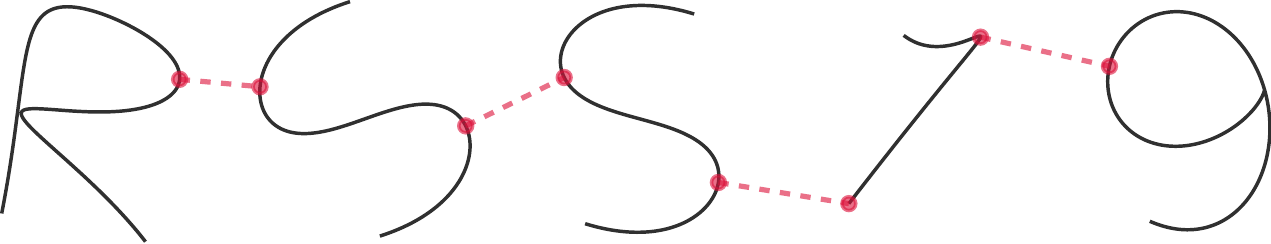}
  \caption{Points closest between pairs of characters 'R', 'S', 'S', '1', and '9' respectively, that are represented as \Bez curves. The evaluation time for the four minimum separating distance queries was 1.14 ms.}
  \label{fig:rss}
\end{figure}

\Cref{alg:mindist_curve} describes the  procedure to evaluate the minimum distance from $\Psi_\mathcal{I}$ to $\Phi_\mathcal{J}$, which is the trace of $\phi: \mathcal{J} \to \mathbb{R}^d$. As one might expect, structurally very little differs between \cref{alg:mindist_curve,alg:mindist}. The collection $\mathcal{L}$ stores pairs of sub-intervals over which the bounds on the solution will be evaluated. The lower and upper bounds are computed using \cref{eq:lowerbound_curve,eq:upperbound_curve} respectively. \Cref{fig:rss} shows the pair of closest points between \Bez curves computed using \cref{alg:mindist_curve}.

\begin{algorithm}[htbp]
    \caption{Minimum Separating Distance $(\Psi_\mathcal{I}, \Phi_\mathcal{J})$} \label{alg:mindist_curve}
    \begin{algorithmic}[1]
        \State $\mathcal{L} \gets \{ \{ \mathcal{I}, \mathcal{J}\}\}$
        \State $\overline{d} \gets \dub(\Psi_\mathcal{I}, \Phi_\mathcal{J})$
        \State $\underline{d}  \gets \dlb(\Psi_\mathcal{I}, \Phi_\mathcal{J})$
        \While{$\overline{d} - \underline{d} > \epsilon$}
            \State $\mathcal{X} \gets \argmin_{\{\mathcal{Q}, \mathcal{R}\} \in \mathcal{L}} \dlb(\Psi_\mathcal{Q}, \Phi_\mathcal{R})$
            \State $\mathcal{X}_L, \mathcal{X}_R \gets \texttt{split}(\mathcal{X})$\label{alg:split_curve}
            \State $\mathcal{L} \gets \mathcal{L} \setminus \{\mathcal{X}\}$
            \State $\mathcal{L} \gets \mathcal{L} \cup \{\mathcal{X}_L, \mathcal{X}_R\}$
            \State $\overline{d} \gets \min_{\{\mathcal{Q}, \mathcal{R}\} \in \mathcal{L}} \dub(\Psi_\mathcal{Q}, \Phi_\mathcal{R})$
            \State $\underline{d}  \gets \min_{\{\mathcal{Q}, \mathcal{R}\} \in \mathcal{L}} \dlb(\Psi_\mathcal{Q}, \Phi_\mathcal{R})$
        \EndWhile
        \State \textbf{return} $\underline{d}$
    \end{algorithmic}
\end{algorithm}

\begin{remark}\label{rem:split}
The $\mathcal{X}_L, \mathcal{X}_R \gets \texttt{split}(\mathcal{X})$ operation from (line~\ref{alg:split}) of \cref{alg:mindist,alg:mindist_curve} returns two mutually disjoint sets such that $\mathcal{X}_L \cup \mathcal{X}_R = \mathcal{X}$. When $\mathcal{X}$ is an interval, the operation must ensure that both $\mathcal{X}_L$ and $\mathcal{X}_R$ are closed. And, when $\mathcal{X}$ is a collection of two intervals, the operation must only split the larger interval so as to preserve the uniform convergence property from \cref{theorem:convergence}. In the implementation\footref{repo}, the $\texttt{split}(\mathcal{X})$ operation bisects the interval $\mathcal{X}$, however, uneven splitting techniques may lead to better performance as seen in \cite{chang2011}.
\end{remark}

\subsection{Tolerance Verification}
As discussed in \cref{sec:intro}, in many applications an exact measurement on the minimum separating distance between objects is not required. A weaker method that verifies if two objects are separated by a distance greater than $\Delta$ may be preferred for computational reasons. Since the states $\overline{d}$ and $\underline{d}$ in \cref{alg:mindist,alg:mindist_curve} keep track of the the bounds on the global optimum, the algorithm will terminate early if either $\overline{d}$ or $\underline{d}$ violate the $\Delta$-tolerance, \ie \ all the convex hulls formed from the curve evaluated at the sub-intervals in collection $\mathcal{L}$ are separated from the object $\mathcal{B}$ by a distance of at least $\Delta$. \Cref{alg:tolver} highlights the differences from \cref{alg:mindist}.

\begin{algorithm}[htbp]
    \caption{Tolerance Verification $(\Psi_\mathcal{I}, B, \Delta)$} \label{alg:tolver}
    \begin{algorithmic}[1]
        \algrenewcommand{\alglinenumber}[1]{}
        \State \vdots
        \algrenewcommand{\alglinenumber}[1]{\footnotesize{4:}}
        \While{$\overline{d} - \Delta > \epsilon$}
            \algrenewcommand{\alglinenumber}[1]{}
            \State \vdots
            \algrenewcommand{\alglinenumber}[1]{\footnotesize{11:}}
            \IfThen{$\underline{d} > \Delta$}{\textbf{return} true}
            \algrenewcommand{\alglinenumber}[1]{\footnotesize{12:}}
        \EndWhile
        \algrenewcommand{\alglinenumber}[1]{\footnotesize{13:}}
        \State \textbf{return} false
    \end{algorithmic}
\end{algorithm}

\subsection{Collision Detection}
Sometimes only the detection of intersection between two objects is required. This is a special case of \cref{alg:tolver}, where $\Delta = 0$. \Cref{alg:tolver} highlights the differences from \cref{alg:mindist}.

\begin{algorithm}[htbp]
    \caption{Collision Detection $(\Psi_\mathcal{I}, B)$} \label{alg:coldet}
    \begin{algorithmic}[1]
        \algrenewcommand{\alglinenumber}[1]{}
        \State \vdots
        \algrenewcommand{\alglinenumber}[1]{\footnotesize{4:}}
        \While{$\overline{d} > \epsilon$}
            \algrenewcommand{\alglinenumber}[1]{}
            \State \vdots
            \algrenewcommand{\alglinenumber}[1]{\footnotesize{11:}}
            \IfThen{$\underline{d} > 0$}{\textbf{return} false}
            \algrenewcommand{\alglinenumber}[1]{\footnotesize{12:}}
        \EndWhile
        \algrenewcommand{\alglinenumber}[1]{\footnotesize{13:}}
        \State \textbf{return} true
    \end{algorithmic}
\end{algorithm}

\section{Computational Results}
\label{sec:results}

We present numerical simulations of our approach for several different proximity query problems. In general, the classes of examples that we consider are categorized as the proximity query between a parametric curve and a point, a convex polygon, and another parametric curve. In these numerical problems we only consider proximity queries in $\mathbb{R}^2$ for ease of viewing, however, the algorithm will hold in general as long as there exist methods to compute $\dlb$ and $\dub$ in the given dimension. Examples of proximity queries for parametric curves in $\mathbb{R}^3$ can be found in the online repository\footref{repo}. In every problem setting, the execution times for computing the minimum separating distance, tolerance verification and collision detection between two objects are benchmarked. Each benchmark result is computed as the median over 10,000 trials of the program. The implementation\footnote{The algorithms are in implemented in Julia \cite{bezanson2017}, and the benchmarks were conducted on a 2.3 GHz Intel Core i5 machine with 8 Gigabytes of RAM.} uses double-precision arithmetic, and we chose the tolerance of $\epsilon = 10^{-10}$ to be used in the optimization procedure. For the tolerance verification queries presented in this section, we choose the value of $\Delta$ as half the minimum separating distance between the objects which they are querying.

\subsection{Curve - Point Proximity}
\begin{figure}[htbp]
  \centering
  \subfloat[]{\includegraphics[width=0.33\columnwidth]{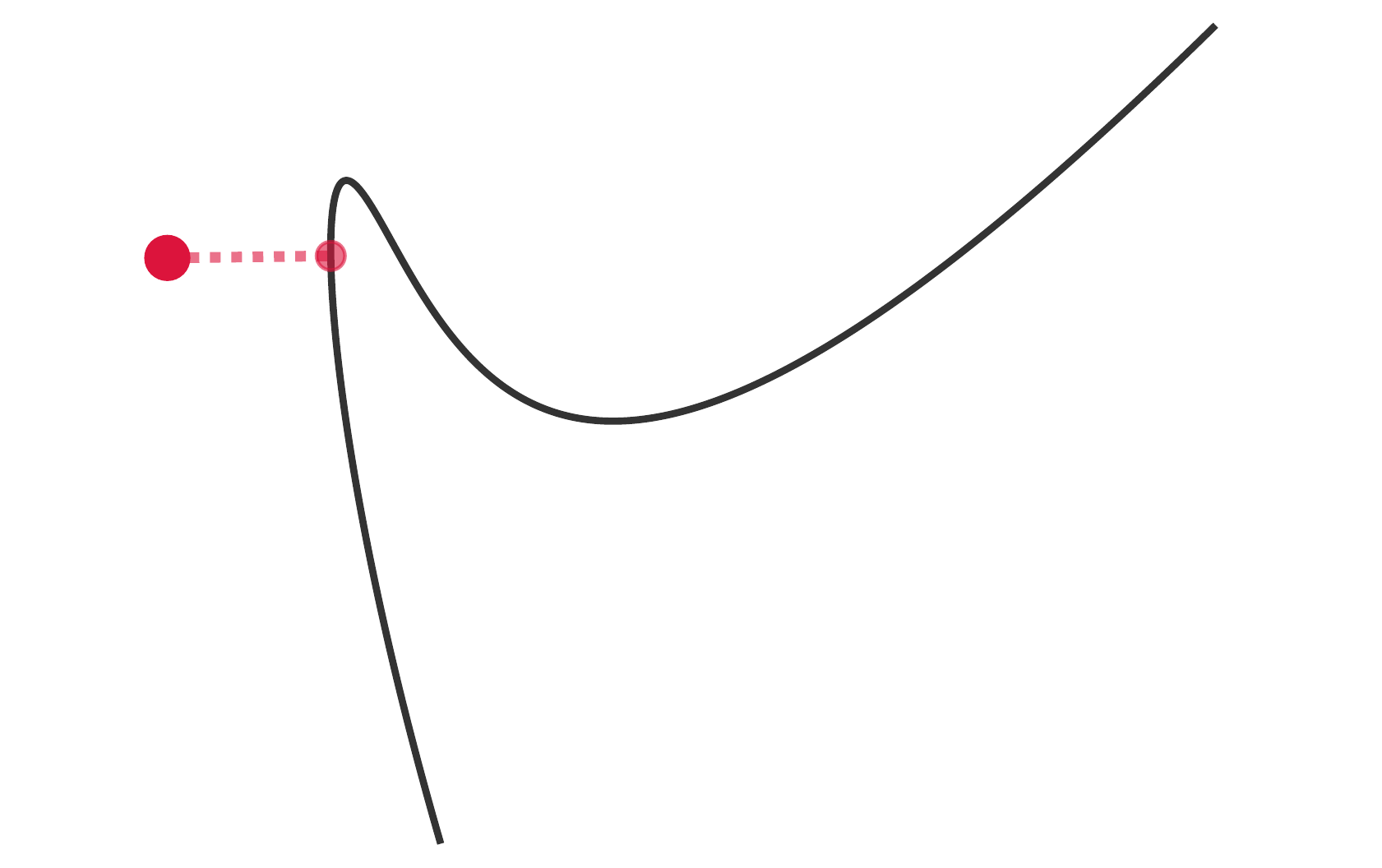}}
  \subfloat[]{\includegraphics[width=0.33\columnwidth]{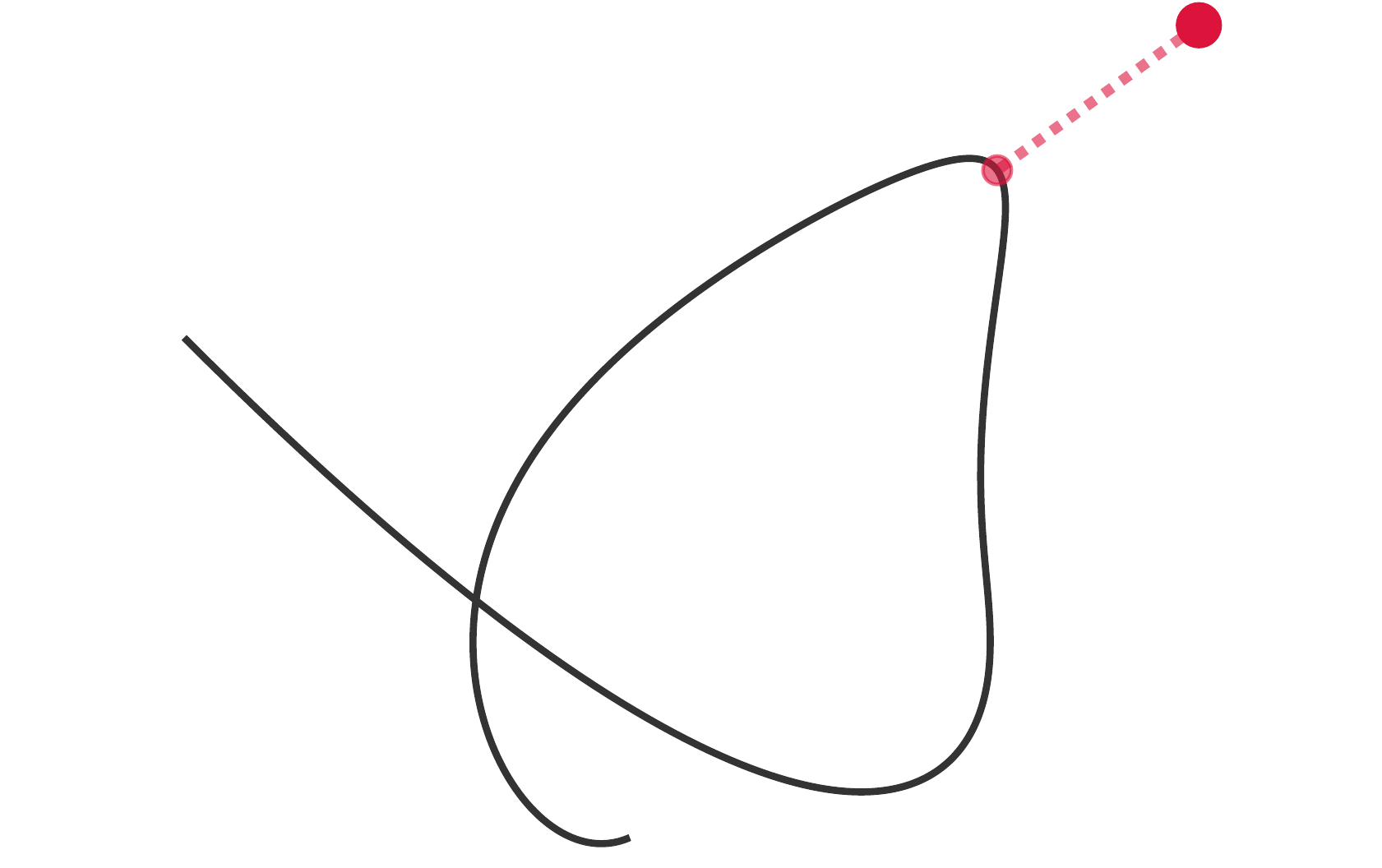}}
  \subfloat[]{\includegraphics[width=0.33\columnwidth]{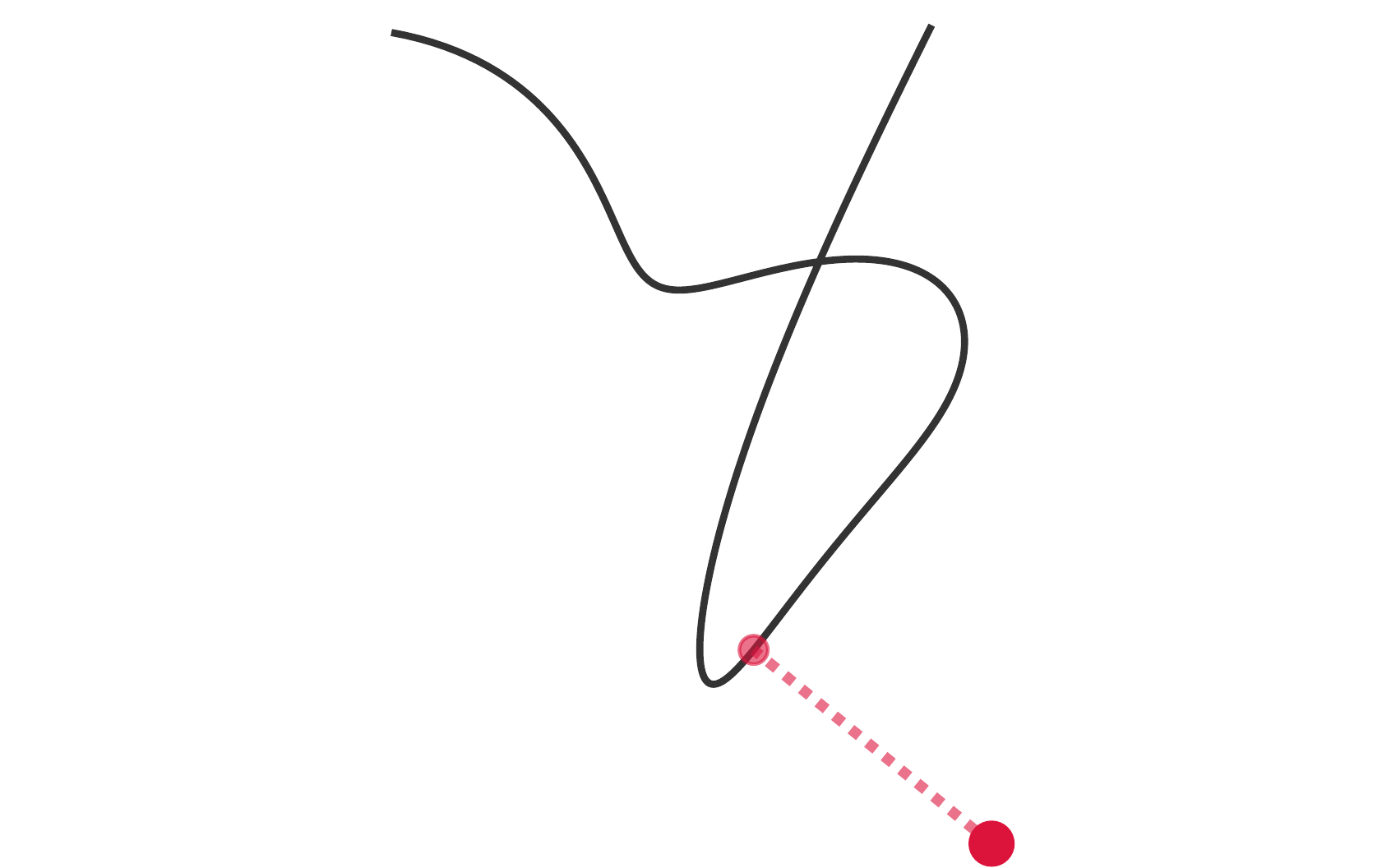}} \\
  \subfloat[]{\includegraphics[width=0.33\columnwidth]{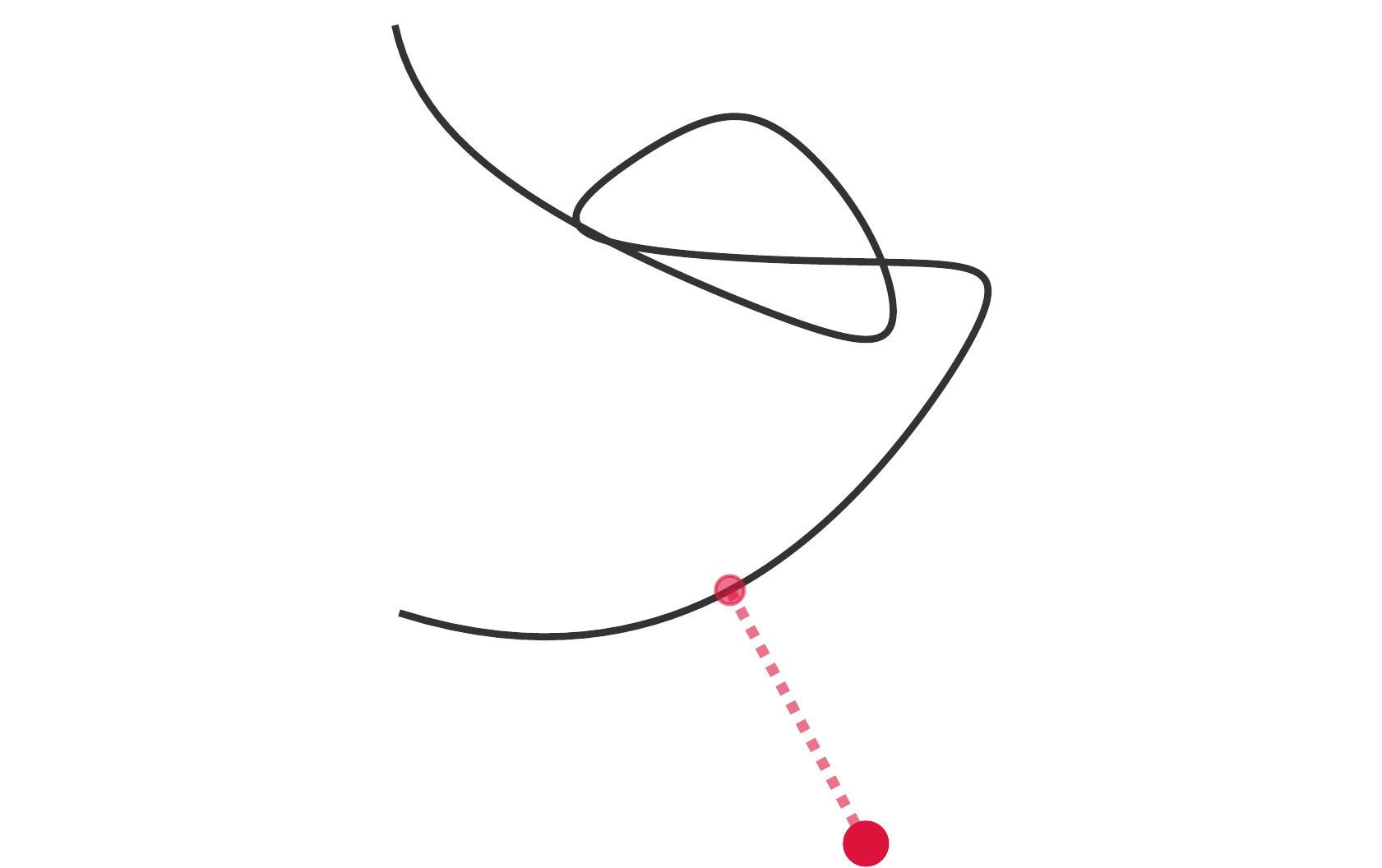}}
  \subfloat[]{\includegraphics[width=0.33\columnwidth]{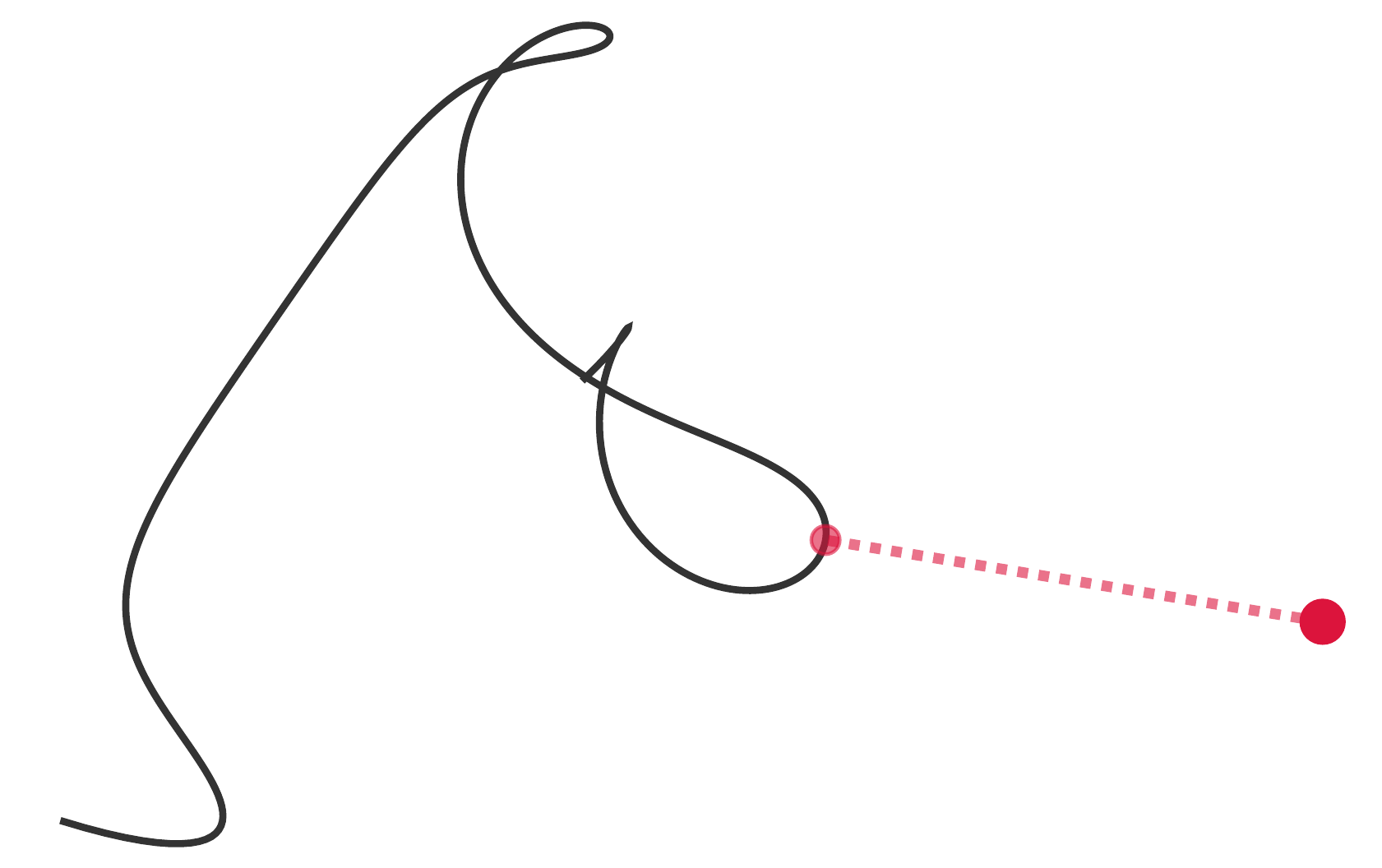}}
  \subfloat[]{\includegraphics[width=0.33\columnwidth]{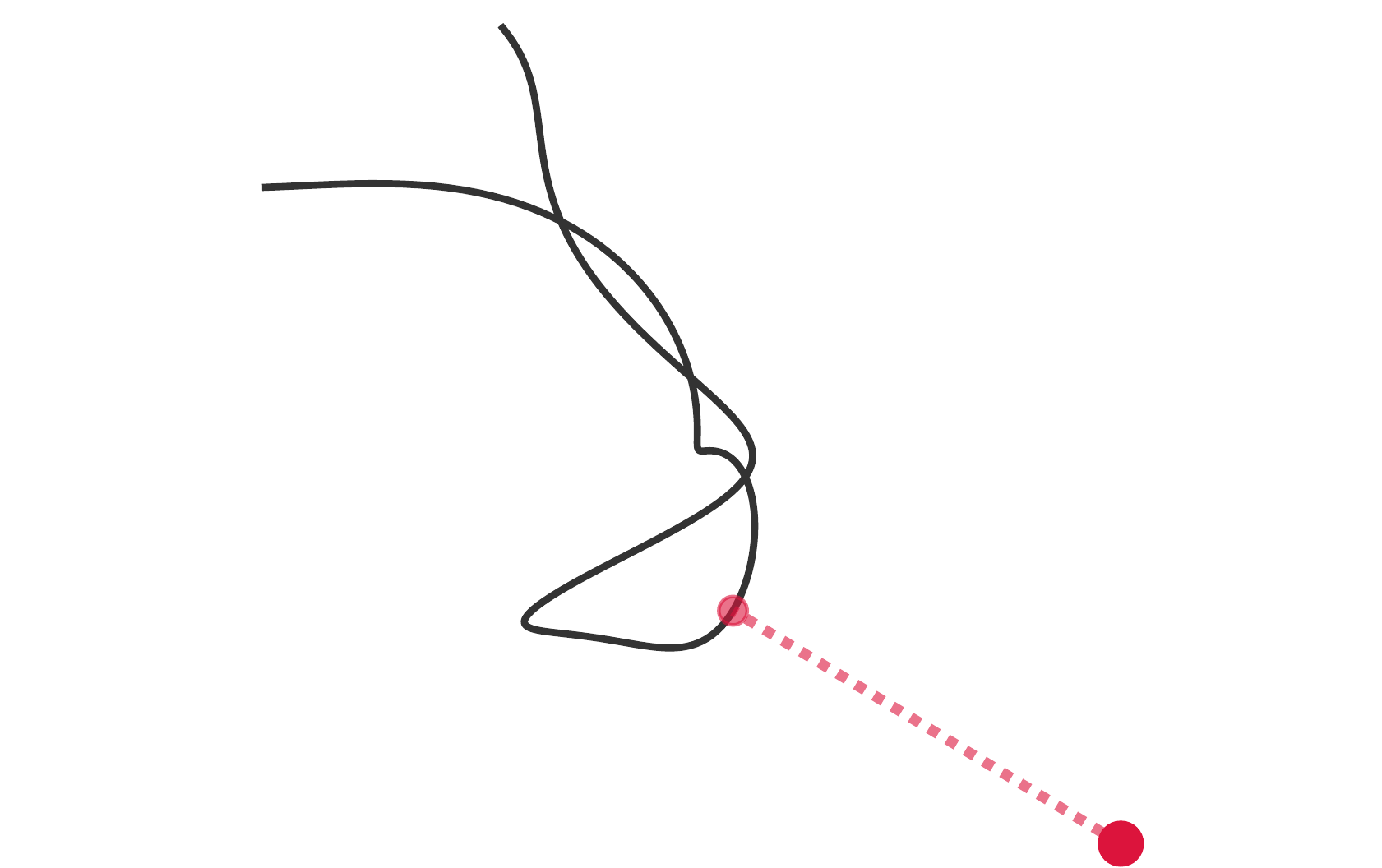}} \\
  \subfloat[]{\includegraphics[width=0.33\columnwidth]{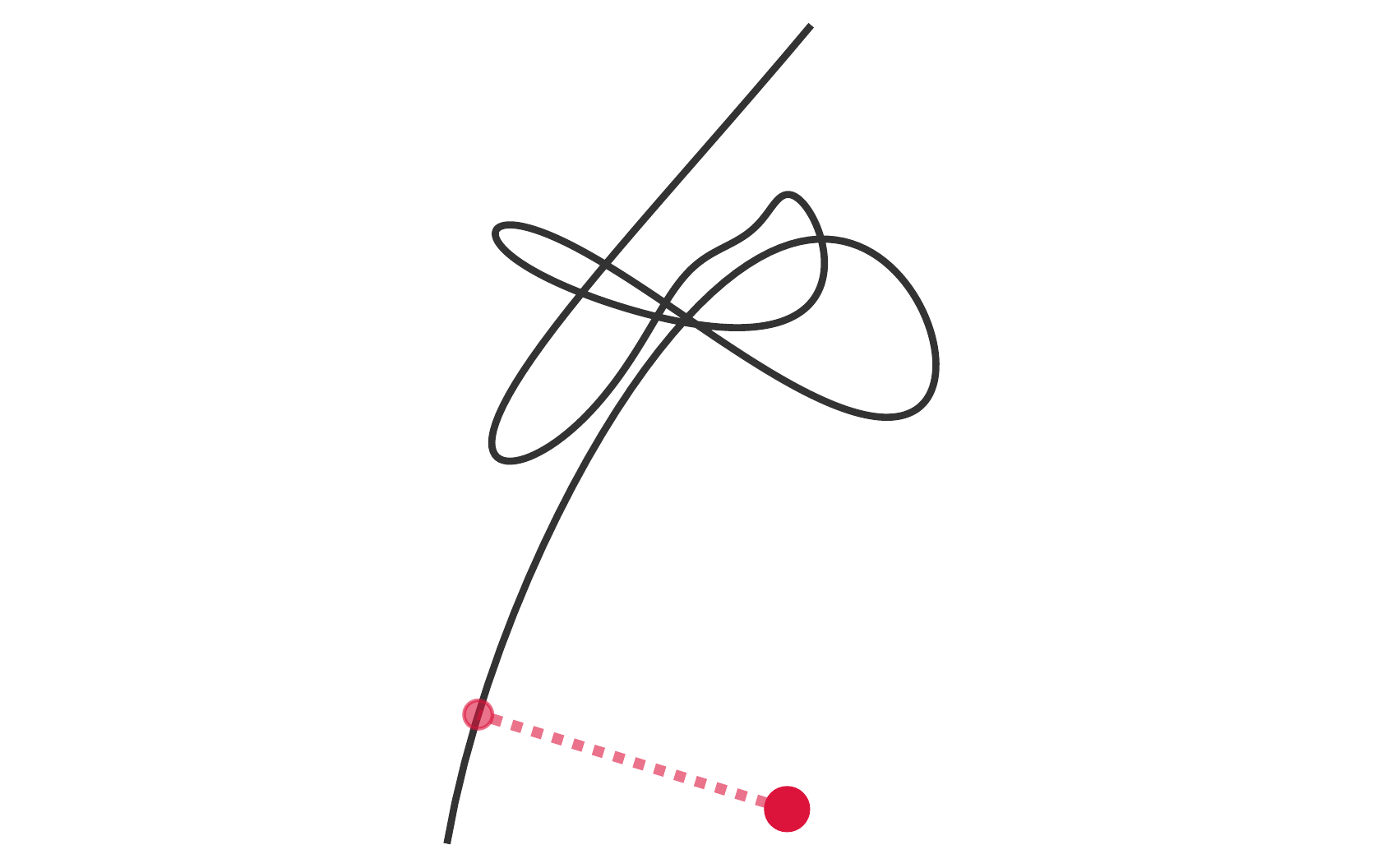}}
  \subfloat[]{\includegraphics[width=0.33\columnwidth]{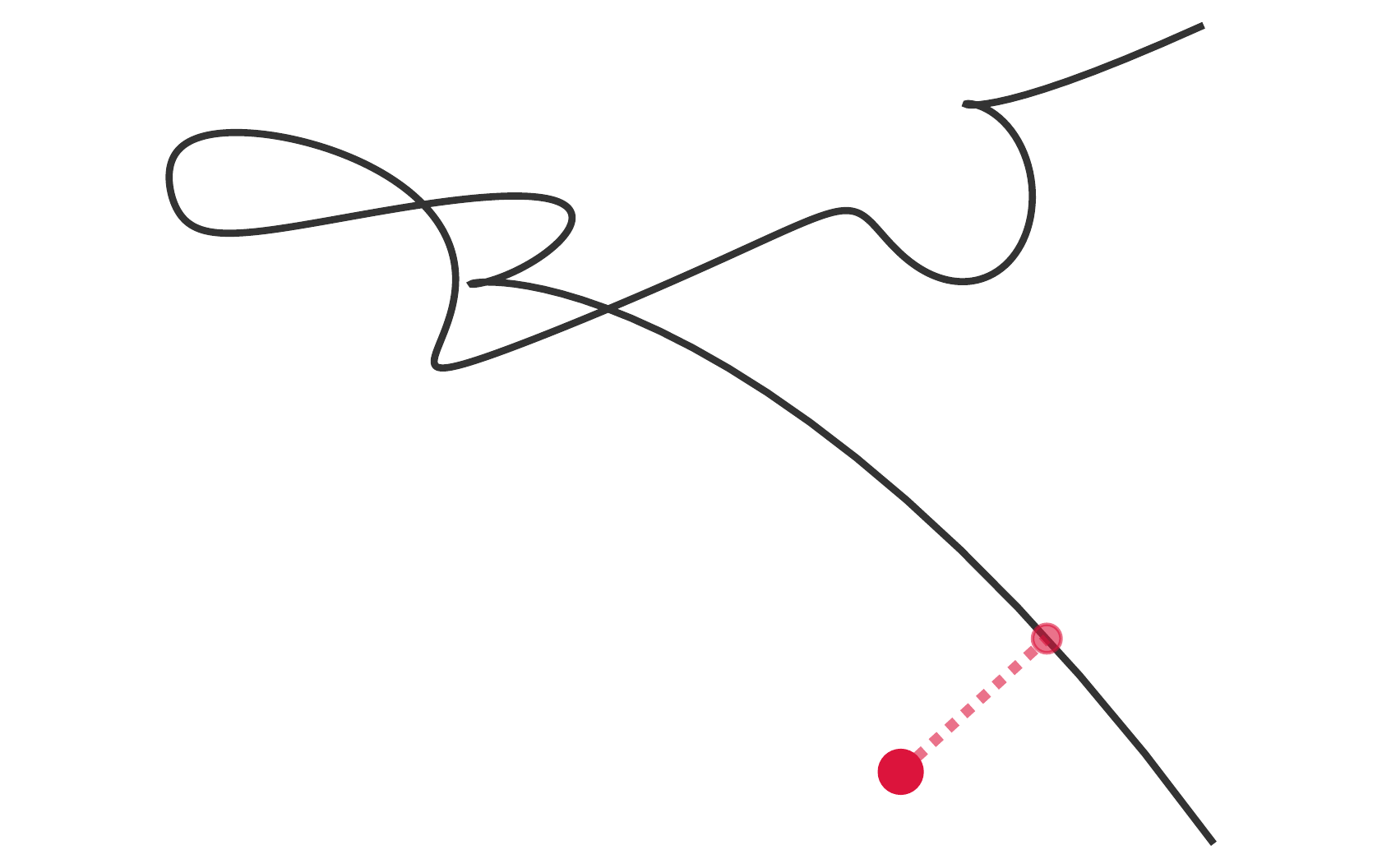}}
  \subfloat[]{\includegraphics[width=0.33\columnwidth]{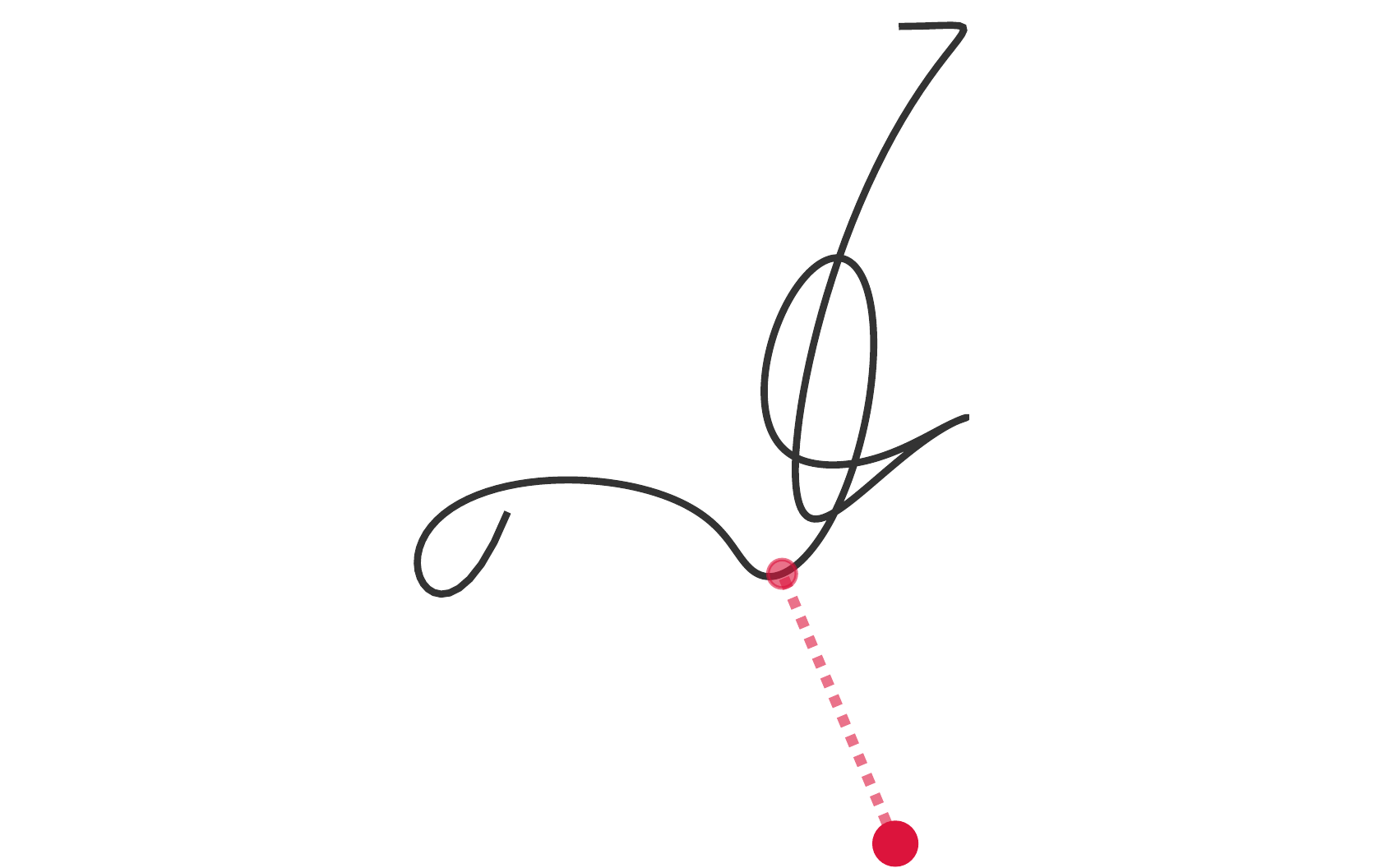}} \\
  \caption{The point on a (a) $\nth{5}$,  (b) $\nth{10}$,  (c) $\nth{15}$,  (d) $\nth{20}$,  (e) $\nth{25}$,  (f) $\nth{30}$,  (g) $\nth{35}$,  (h) $\nth{40}$, and (h) $\nth{45}$  \Bez curve that is closest to another randomly chosen point.}
  \label{fig:curvept}
\end{figure}

We show the performance of computing proximity queries between \Bez curves and a point in $\mathbb{R}^2$ using our approaches and the approach found in \cite{chang2011}. The control points of the curve are uniformly randomly placed in $[0,1] \times [0,1]$, and the point to which the proximity is computed is also randomly placed in the same unit square. \Cref{fig:curvept} shows a few examples of the problems that were considered.

\begin{figure}[htbp]
  \centering
  \subfloat[]{\label{fig:compare_time}\includegraphics[width=0.5\columnwidth]{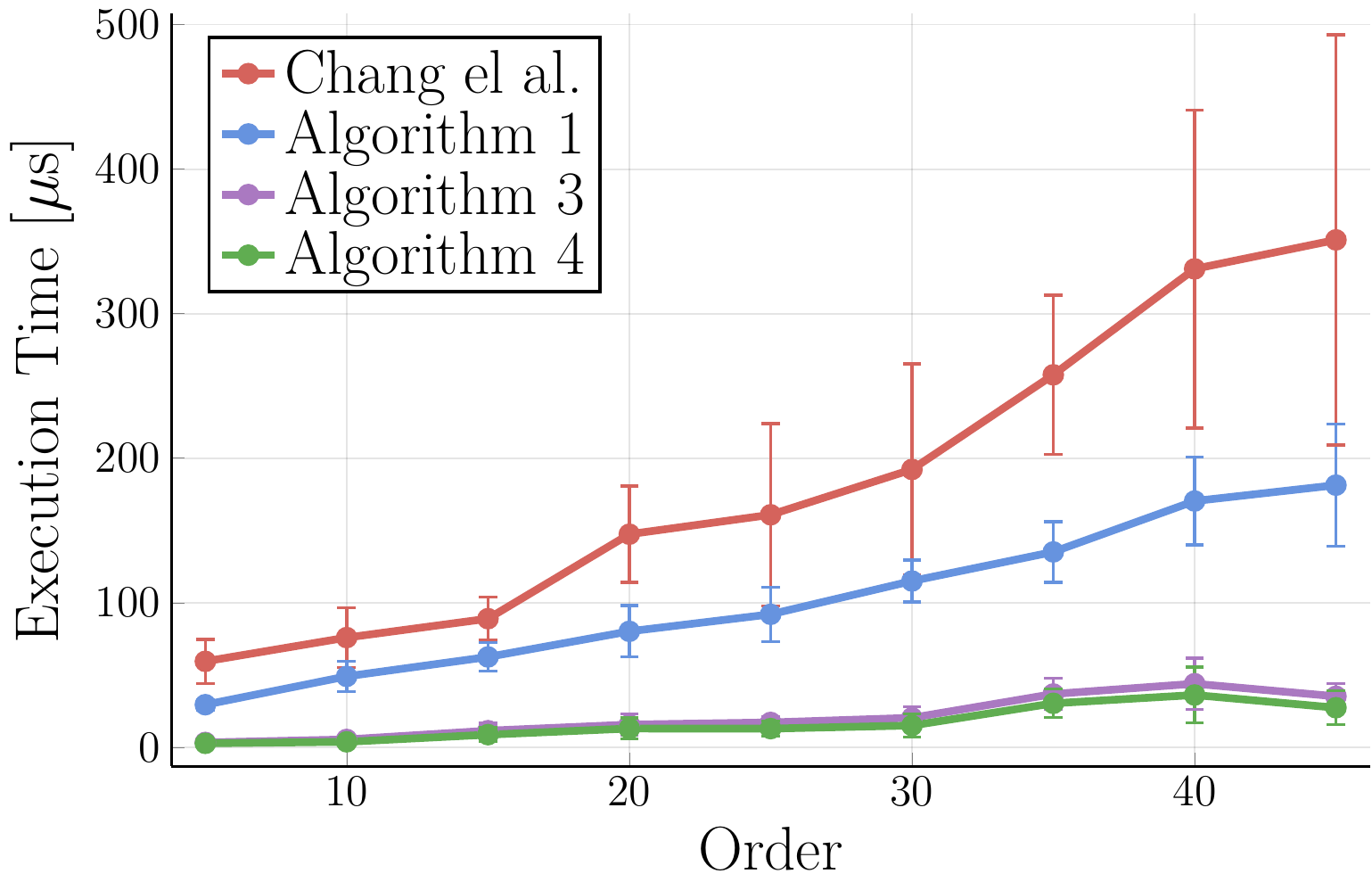}}
  \subfloat[]{\label{fig:compare_mem}\includegraphics[width=0.5\columnwidth]{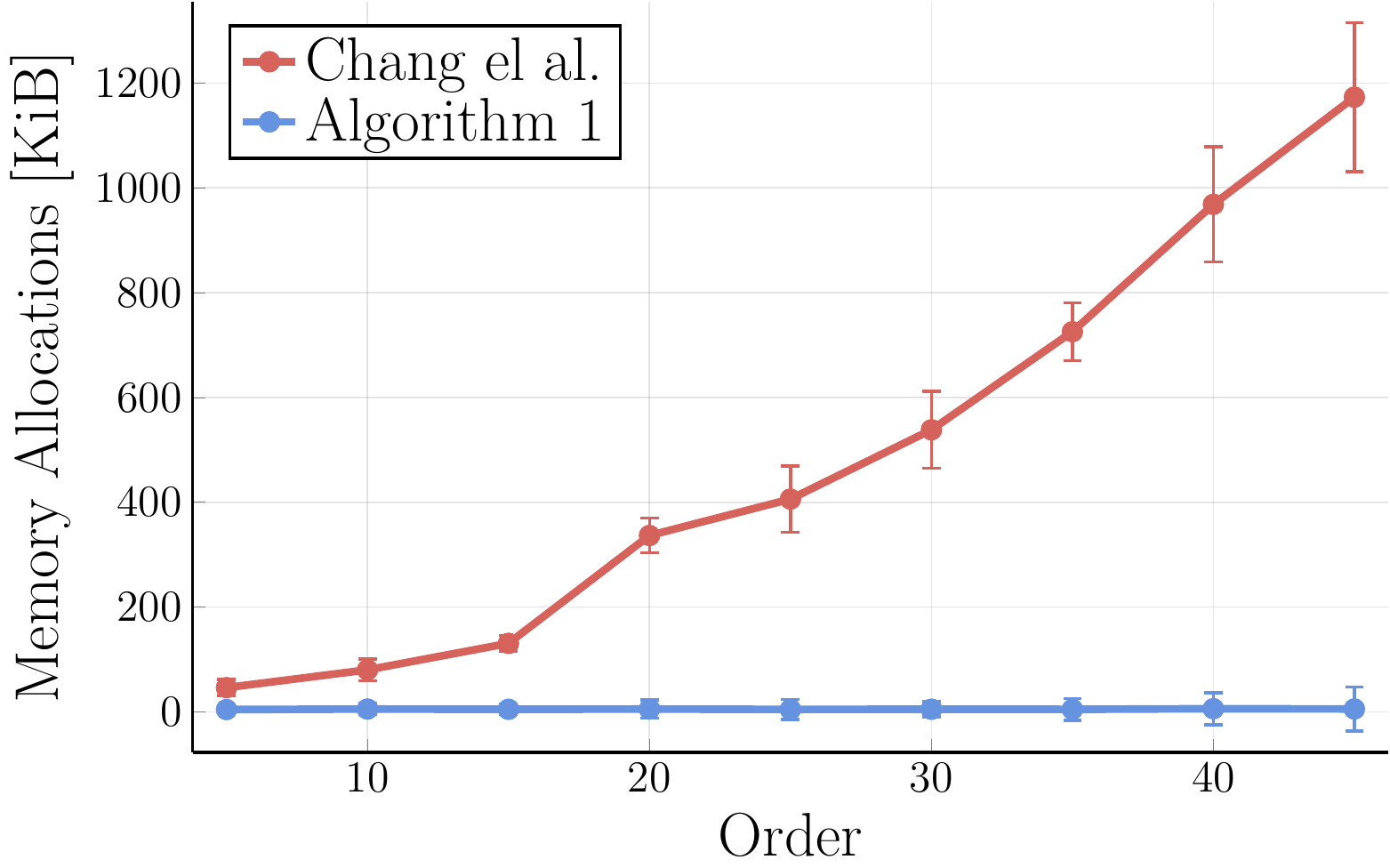}}
  \caption{The average computational time (a) and allocated memory (b) to evaluate a proximity query between ten randomly generated \Bez curves and a point. The algorithm presented in~\cite{chang2011} and \cref{alg:mindist} both use a tolerance $\epsilon = 10^{-10}$. The error bars represent the standard deviation of the trial runs.}
  \label{fig:compare}
\end{figure}

The numerical simulations compute the mean execution time over a range of different order \Bez curves as shown in \cref{fig:compare}. Notice that the computational efficiency (\cref{fig:compare_time}) of \cref{alg:mindist} is improved compared to that of the curve subdivision algorithm in \cite{chang2011}. This is in part because the construction of the bounding region in \cite{chang2011} is the control polygon of the subdivided curve that is obtained through the expensive De Casteljau's algorithm \cite{farin2000}. Furthermore, the parameterizations of the subdivided curves are stored in a queue, which is very memory expensive for higher order curves as seen in \cref{fig:compare_mem}. On the other hand, our methods are very memory efficient as the priority queue (expressed as the collection $\mathcal{L}$ in \cref{alg:mindist} only contains information of the intervals.   Additionally, computing proximity queries for \Bez curves using our approach is numerically robust as \cref{eq:arclength_ub} is also a \Bez curve and does not require any change of basis. Approaches to compute the closed-form expression of \cref{eq:arclength_ub} can be found in \cite{farouki2012}. 

Both \cref{alg:tolver,alg:coldet} significantly outperform the other methods. This is a result of the fact that the algorithms are terminated as soon as the bounds are met on the $\Delta$-tolerance, rather than proceeding to obtain an $\epsilon$-suboptimal solution.

\begin{figure*}[htbp]
  \centering
  \subfloat[]{\label{fig:sim_heart}\includegraphics[width=0.33\columnwidth]{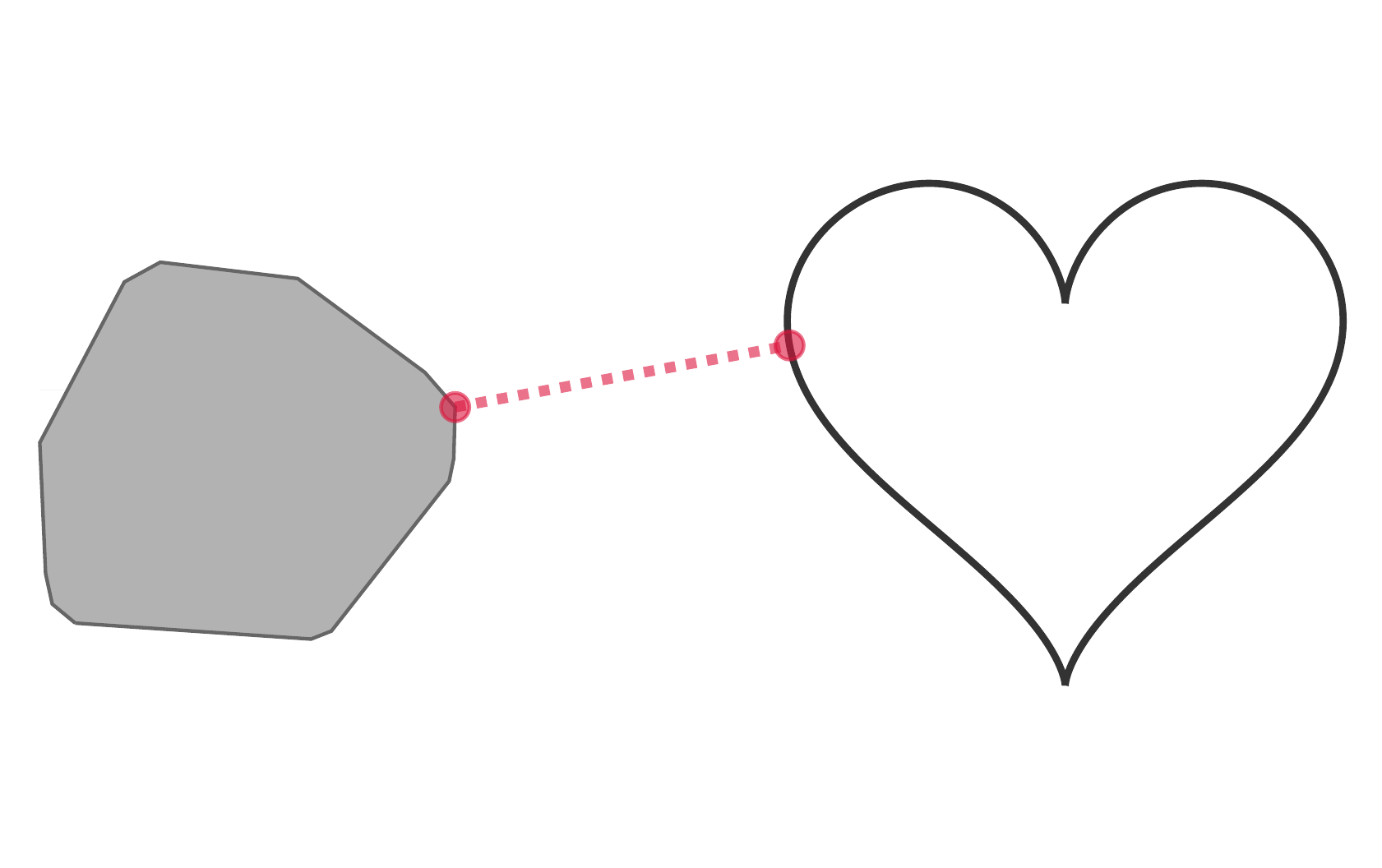}}
  \subfloat[]{\label{fig:sim_ranunculoid}\includegraphics[width=0.33\columnwidth]{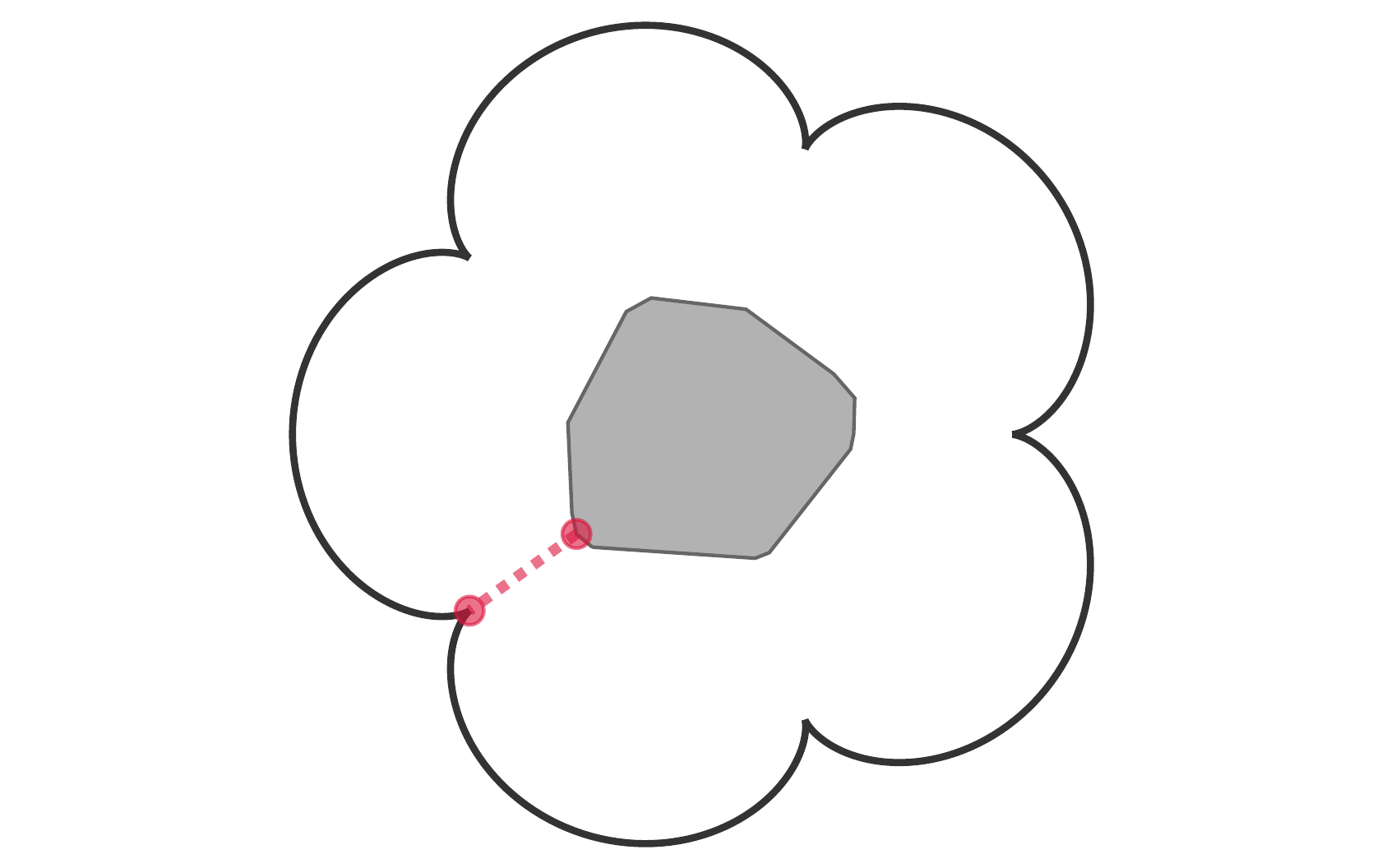}}
  \subfloat[]{\label{fig:sim_clothoid}\includegraphics[width=0.33\columnwidth]{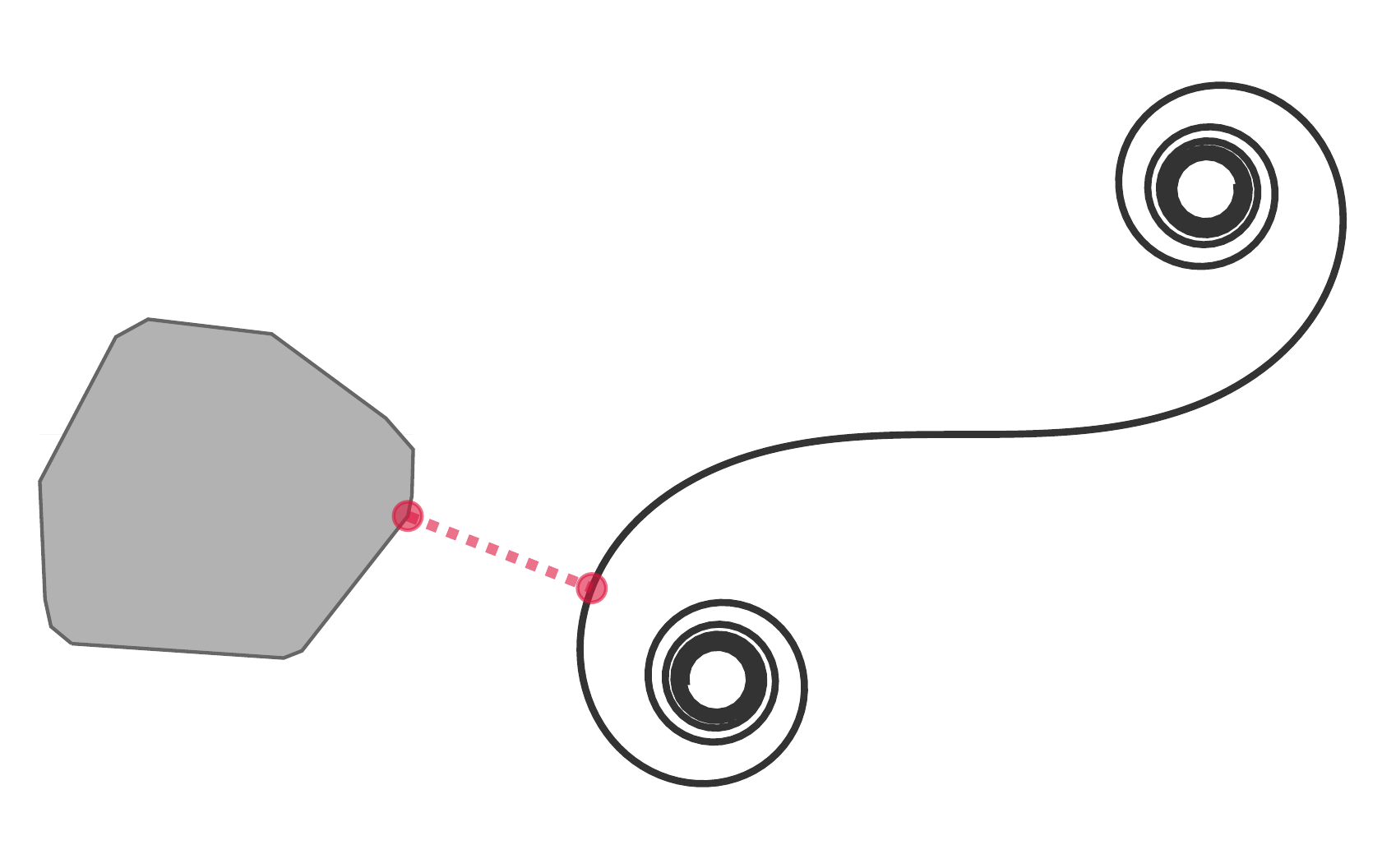}}
  \subfloat[]{\label{fig:sim_bez_10}\includegraphics[width=0.33\columnwidth]{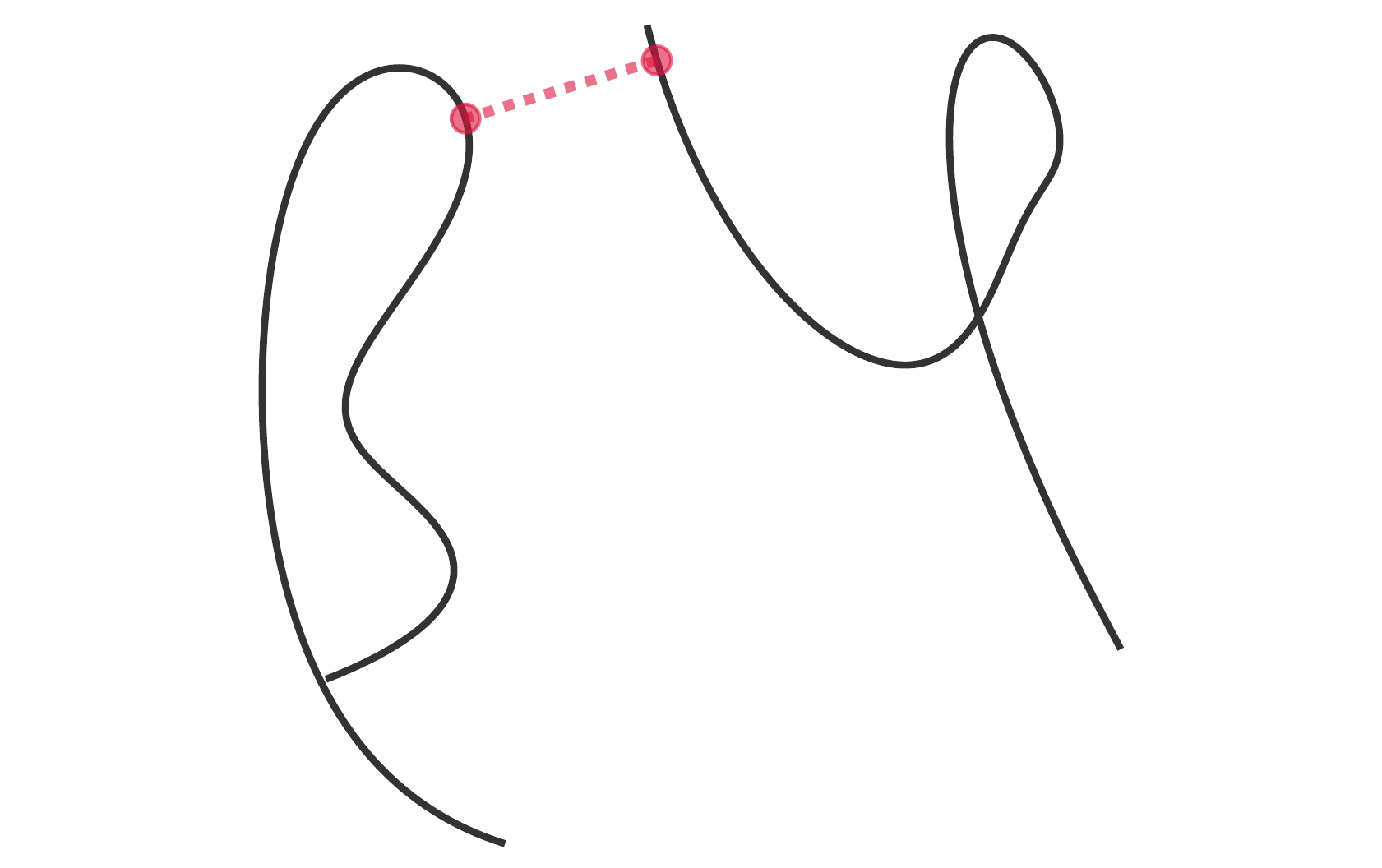}}
  \subfloat[]{\label{fig:sim_circ_circ}\includegraphics[width=0.33\columnwidth]{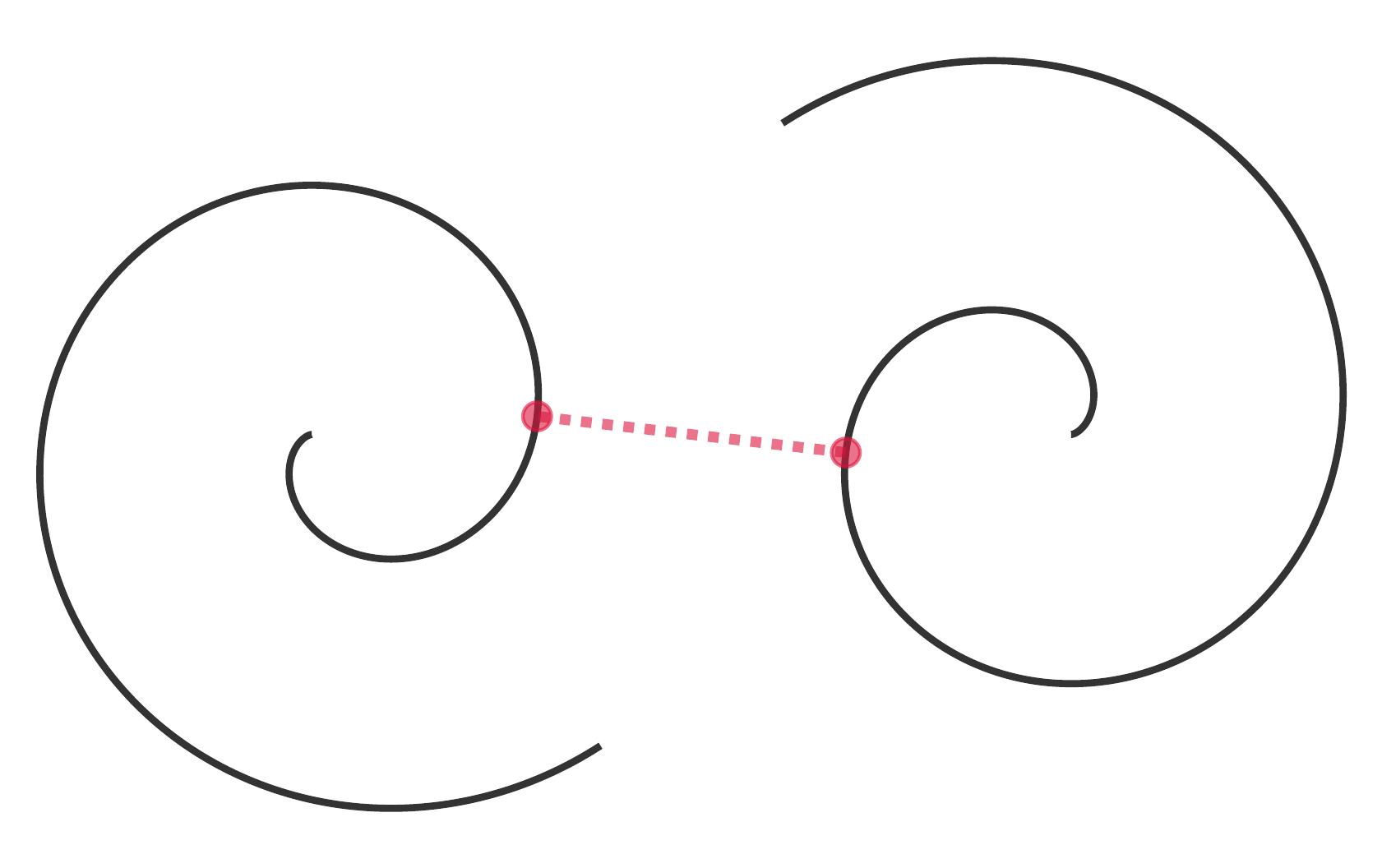}}
  \subfloat[]{\label{fig:sim_liss_fish}\includegraphics[width=0.33\columnwidth]{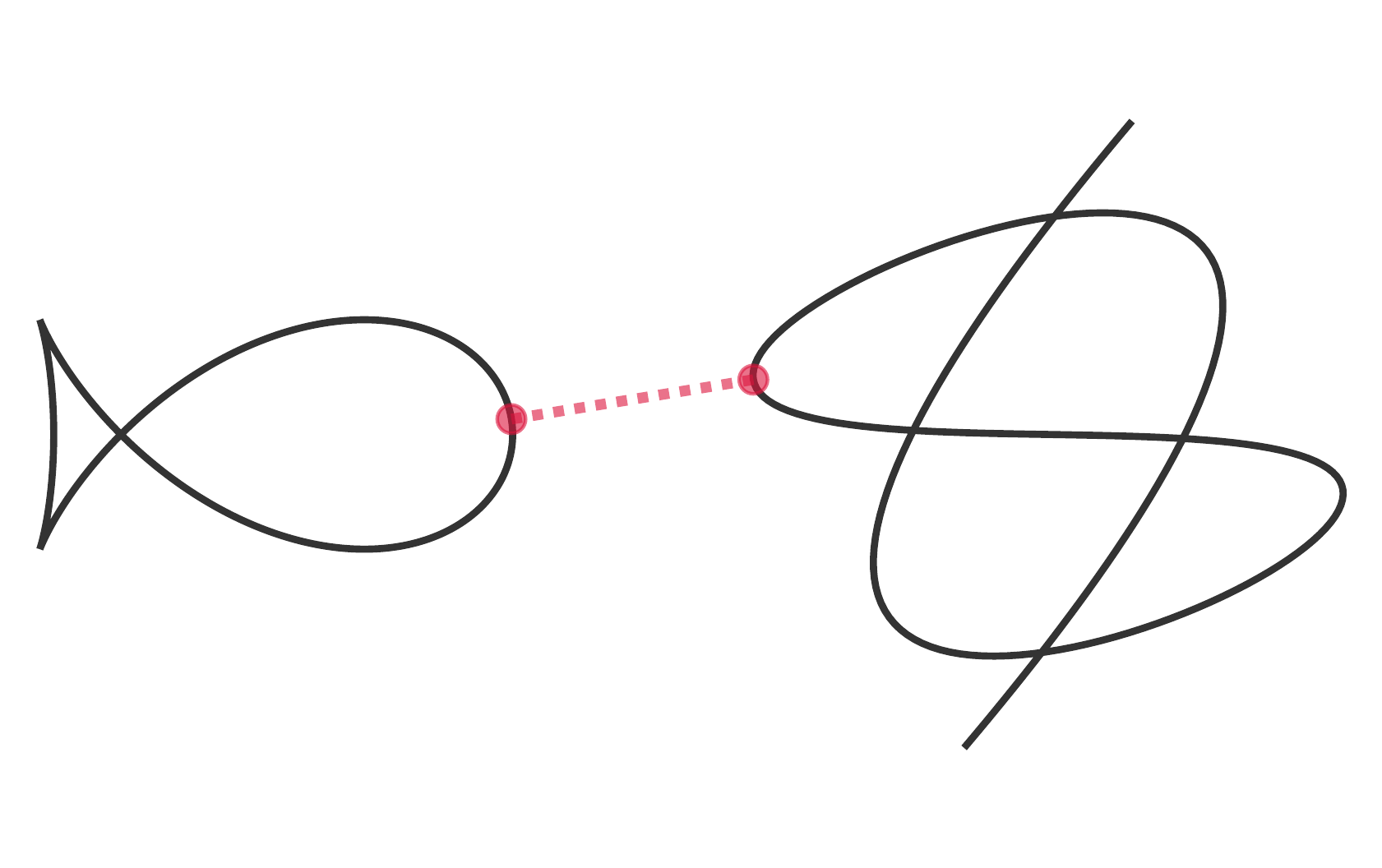}}
  \caption{Points closest between (a) a convex polygon and a heart-shaped curve \cite{weisstein2003heart}, (b) a convex polygon and an epicycloid \cite{weisstein2003ranunculoid}, (c) a convex polygon and an Euler spiral, (d) two $\nth{10}$ order \Bez curves, (e) two involutes of a circle, and (f) a fish curve \cite{weisstein2003fish} and a Lissajous curve.}
  \label{fig:curvecvx}
\end{figure*}

\begin{table*}[htbp]
    \small
    \centering
    \begin{tabular}{  c||cccccc}
       \toprule \toprule
        \multicolumn{1}{c||}{\multirow{2}{*}{\textbf{Query}}} & \multicolumn{6}{c}{\textbf{Median Time} ($\mu$s)}  \\
        \\[-1em]
        \cline{2-7}\\[-1em]
        & (a) & (b) & (c) & (d) & (e) & (f)\\
        \hline \hline \\[-1em]
        Alg. 1/2 & 208.22 & 234.52 &  150.03 $\times 10^3$ & 425.66 &  1.21 $\times 10^3$ & 3.74 $\times 10^3$\\
        Alg. 3 & 10.52 & 80.90 &  29.70 $\times 10^3$ & 49.20 & 204.68 & 471.99\\
        Alg. 4 & 10.38 & 59.19 &  17.08 $\times 10^3$ & 37.66 & 116.22 & 229.25\\
        \toprule
    \end{tabular}
    \caption{The median computational time to evaluate a proximity query between each of the objects in \cref{fig:curvecvx}.}
    \label{tab:curvecvx}
\end{table*}

\subsection{Curve - Convex Polygon Proximity}
We proceed to discuss the performance of proximity queries between parametric curves and polygons through some examples in \cref{fig:curvecvx}. We consider the general class of absolutely continuous parametric curves, and, since not all the curves can be transformed\footnote{Even transformations between polynomial bases are numerically unstable for higher order polynomials \cite{hermann1996}.} to a Bernstein basis,  we do not provide numerical comparisons to \cite{chang2011} here. For the underlying routines that compute the bounds on the solution, \ie \ $\dlb$ and $\dub$, we use the GJK-algorithm \cite{gilbert1988} to compute the distances between the polygons and the ellipsoidal convex hulls. \Cref{fig:sim_heart,fig:sim_ranunculoid,fig:sim_clothoid} show proximity of curves with a sinusoidal basis with convex polygons. It should be noted that although the Euler spiral is defined over $\mathbb{R}$, we consider its trace \cref{fig:sim_clothoid} in the compact subset: $[-2\pi, 2\pi]$. In \cref{tab:curvecvx}, the median execution times for the different problems are enumerated. It should be highlighted that the reason for the longer run times of the Euler spiral is due to the evaluation of the parametric function that requires numerical integration.

\subsection{Curve - Curve Proximity}
Similar to above, \cref{tab:curvecvx} also shows the performance for computing the proximity queries for the problems in \cref{fig:sim_bez_10,fig:sim_circ_circ,fig:sim_liss_fish}. As these problems require simultaneously searching across dual parameter spaces, it is natural that the proximity queries will have longer run times. This will be offset if a lower $\epsilon$-tolerance is chosen in the procedures.

\subsection{Trajectory Replanning Example}
\begin{figure}[htbp]
  \centering
  \includegraphics[width=\columnwidth]{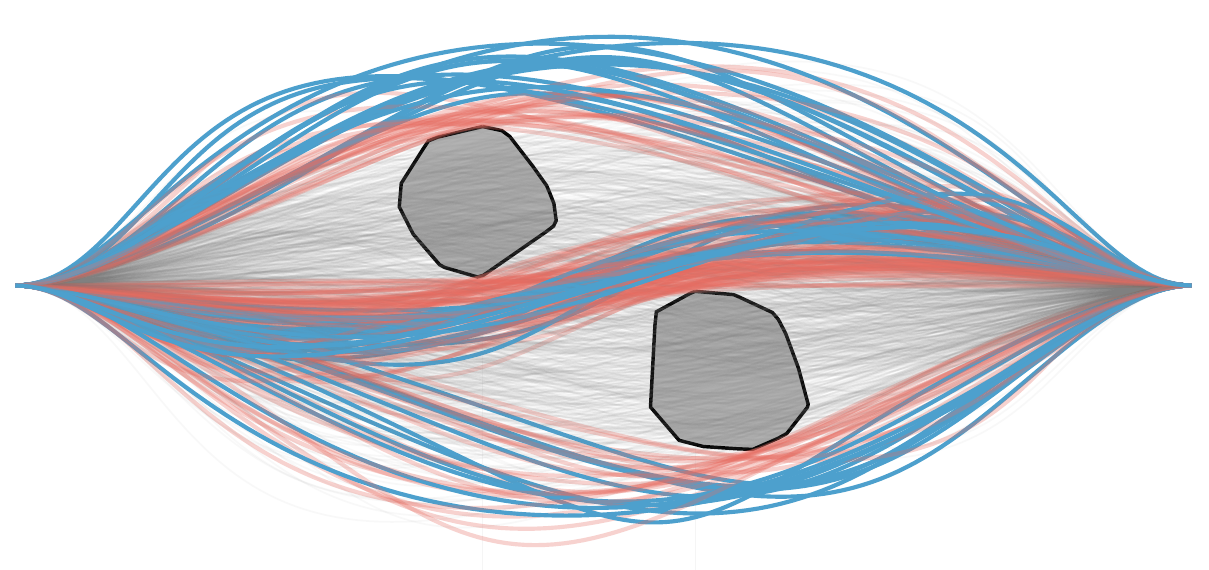}
  \caption{An example scenario of a robot (starting on the left) attempting to replan its trajectory around obstacles: The blue trajectories represent feasible solutions, the red trajectories are collision-free, but infeasible because of their close proximity to the obstacles (\ie $\le \Delta$), and the grey trajectories are infeasible, because they collide with the obstacles in the environment.}
  \label{fig:obstacle_avoid}
\end{figure}

To demonstrate the benefit of fast collision detection, Figure~\ref{fig:obstacle_avoid} shows a common situation in path replanning problems. A vehicle is confronted with an obstacle and randomly samples a large number of possible trajectories from a distribution to plan a path around the obstacle \cite{kobilarov2012}. For each of these trajectories, a certificate of collision avoidance and minimum safety distance are necessary. In this situation, 1000 quintic \Bez curves are generated and, using \cref{alg:tolver}, a tolerance verification query is performed against both the obstacles with a total run time of 11.61 milliseconds. In the simulation, 818 trajectories of the 1000 samples are in collision with at least one of the two obstacles, 116 trajectories are collision free but violate the minimum safety distance constraint, and only the remaining 66 trajectories are feasible. This ability to rapidly compute the feasibility of trajectories allow for a large sample size of trajectories to be validated in a very short amount of time. In addition, these methods prove beneficial for predicting collisions in dynamic environments wherein only probabilistic information of the obstacle behavior is available \cite{patterson2019} by computing proximity queries with the boundary of the confidence region of an obstacle's trajectory. 

\section{Conclusions}
We have presented new algorithms for proximity queries evaluating minimum distance, tolerance verification and collision detection. To the authors' knowledge, this is the first proximity query method that works on such a large class of parametric curves. In addition, the methods show improved computational speed, even compared to the methods taking advantage of a known curve basis. 

The presented proximity query algorithms are built around an interval branch-and-bound method that provide $\epsilon$-suboptimal solutions. We present efficient methods to construct convex hulls that enclose a parametric curve based on the upper bound of its arc length. The convex hulls are used to compute the lower bound on the minimum separating distance to obstacles during the execution of the algorithm. Using this approach, computational results are shown for the evaluation of minimum distance, tolerance verification and collision detection between arbitrary absolutely continuous parametric curves.


\section*{Acknowledgments}
This work was developed with financial support from the National Science Foundation's National Robotics Initiative (\#1830639, \#1528036), Office of Naval Research (N00014-19-1-2106), National Aeronautics and Space Administration (NASA), and the Air Force Office of Scientific Research.


\bibliographystyle{plainnat}
\bibliography{main.bib}

\begin{thebibliography}{39}
\providecommand{\natexlab}[1]{#1}
\providecommand{\url}[1]{\texttt{#1}}
\expandafter\ifx\csname urlstyle\endcsname\relax
  \providecommand{\doi}[1]{doi: #1}\else
  \providecommand{\doi}{doi: \begingroup \urlstyle{rm}\Url}\fi

\bibitem[Barto{\v{n}} and Elber(2011)]{bartovn2011}
Michael Barto{\v{n}} and Gershon Elber.
\newblock {Spiral fat arcs--Bounding regions with cubic convergence}.
\newblock \emph{Graphical Models}, 73\penalty0 (2):\penalty0 50--57, 2011.

\bibitem[Bezanson et~al.(2017)Bezanson, Edelman, Karpinski, and
  Shah]{bezanson2017}
Jeff Bezanson, Alan Edelman, Stefan Karpinski, and Viral~B Shah.
\newblock Julia: A fresh approach to numerical computing.
\newblock \emph{SIAM review}, 59\penalty0 (1):\penalty0 65--98, 2017.

\bibitem[Boyd and Mattingley(2007)]{boyd2007}
Stephen Boyd and Jacob Mattingley.
\newblock {Branch and bound methods}.
\newblock 2007.

\bibitem[Chakraborty et~al.(2008)Chakraborty, Peng, Akella, and
  Mitchell]{chakraborty2008}
Nilanjan Chakraborty, Jufeng Peng, Srinivas Akella, and John~E Mitchell.
\newblock {Proximity queries between convex objects: An interior point approach
  for implicit surfaces}.
\newblock \emph{IEEE Transactions on Robotics}, 24\penalty0 (1):\penalty0
  211--220, 2008.

\bibitem[Chang et~al.(2011)Chang, Choi, Kim, and Wang]{chang2011}
Jung-Woo Chang, Yi-King Choi, Myung-Soo Kim, and Wenping Wang.
\newblock {Computation of the minimum distance between two {B\'ezier}
  curves/surfaces}.
\newblock \emph{Computers \& Graphics}, 35\penalty0 (3):\penalty0 677--684,
  2011.

\bibitem[Chen et~al.(2009)Chen, Chen, Wang, Xu, Yong, and Paul]{chen2009}
Xiao-Diao Chen, Linqiang Chen, Yigang Wang, Gang Xu, Jun-Hai Yong, and
  Jean-Claude Paul.
\newblock {Computing the minimum distance between two {B\'ezier} curves}.
\newblock \emph{Journal of Computational and Applied Mathematics}, 229\penalty0
  (1):\penalty0 294--301, 2009.

\bibitem[Cichella et~al.(2018{\natexlab{a}})Cichella, Kaminer, Walton, and
  Hovakimyan]{cichella2018}
Venanzio Cichella, Isaac Kaminer, Claire Walton, and Naira Hovakimyan.
\newblock {Optimal motion planning for differentially flat systems using
  {Bernstein} approximation}.
\newblock \emph{IEEE Control Systems Letters}, 2\penalty0 (1):\penalty0
  181--186, 2018{\natexlab{a}}.

\bibitem[Cichella et~al.(2018{\natexlab{b}})Cichella, Kaminer, Walton,
  Hovakimyan, and Pascoal]{cichella2018b}
Venanzio Cichella, Isaac Kaminer, Claire Walton, Naira Hovakimyan, and Antonio
  Pascoal.
\newblock Bernstein approximation of optimal control problems.
\newblock \emph{arXiv preprint arXiv:1812.06132}, 2018{\natexlab{b}}.

\bibitem[Ehmann and Lin(2001)]{ehmann2001}
Stephen~A Ehmann and Ming~C Lin.
\newblock {Accurate and fast proximity queries between polyhedra using convex
  surface decomposition}.
\newblock In \emph{Computer Graphics Forum}, volume~20, pages 500--511. Wiley
  Online Library, 2001.

\bibitem[Elber and Grandine(2008)]{elber2008}
Gershon Elber and Tom Grandine.
\newblock Hausdorff and minimal distances between parametric freeforms in
  {$\mathbb{R}^2$} and {$\mathbb{R}^3$}.
\newblock In \emph{International Conference on Geometric Modeling and
  Processing}, pages 191--204. Springer, 2008.

\bibitem[Farin and Hansford(2000)]{farin2000}
Gerald Farin and Dianne Hansford.
\newblock \emph{The essentials of {CAGD}}.
\newblock AK Peters/CRC Press, 2000.

\bibitem[Farouki(2012)]{farouki2012}
Rida~T Farouki.
\newblock {The Bernstein polynomial basis: A centennial retrospective}.
\newblock \emph{Computer Aided Geometric Design}, 29\penalty0 (6):\penalty0
  379--419, 2012.

\bibitem[Gilbert et~al.(1988)Gilbert, Johnson, and Keerthi]{gilbert1988}
Elmer~G Gilbert, Daniel~W Johnson, and S~Sathiya Keerthi.
\newblock {A fast procedure for computing the distance between complex objects
  in three-dimensional space}.
\newblock \emph{IEEE Journal on Robotics and Automation}, 4\penalty0
  (2):\penalty0 193--203, 1988.

\bibitem[Gottschalk et~al.(1996)Gottschalk, Lin, and Manocha]{gottschalk1996}
Stefan Gottschalk, Ming~C Lin, and Dinesh Manocha.
\newblock {{OBBTree}: A hierarchical structure for rapid interference
  detection}.
\newblock In \emph{Proceedings of the 23rd annual conference on Computer
  graphics and interactive techniques}, pages 171--180. ACM, 1996.

\bibitem[Hermann(1996)]{hermann1996}
Thomas Hermann.
\newblock {On the stability of polynomial transformations between {Taylor,
  Bernstein and Hermite} forms}.
\newblock \emph{Numerical algorithms}, 13\penalty0 (2):\penalty0 307--320,
  1996.

\bibitem[Iseki(1960)]{iseki1960}
Kanesiroo Iseki.
\newblock {On certain properties of parametric curves}.
\newblock \emph{Journal of the Mathematical Society of Japan}, 12\penalty0
  (2):\penalty0 129--173, 1960.

\bibitem[Jaulin et~al.(2001)Jaulin, Kieffer, Didrit, and Walter]{jaulin2001}
Luc Jaulin, Michel Kieffer, Olivier Didrit, and Eric Walter.
\newblock \emph{{Applied interval analysis: With examples in parameter and
  state estimation, robust control and robotics}}, volume~1.
\newblock Springer Science \& Business Media, 2001.

\bibitem[Kleinbort et~al.(2016)Kleinbort, Salzman, and Halperin]{kleinbort2016}
Michal Kleinbort, Oren Salzman, and Dan Halperin.
\newblock {Collision detection or nearest-neighbor search? On the computational
  bottleneck in sampling-based motion planning}.
\newblock \emph{arXiv preprint arXiv:1607.04800}, 2016.

\bibitem[Kobilarov(2012)]{kobilarov2012}
Marin Kobilarov.
\newblock Cross-entropy motion planning.
\newblock \emph{The International Journal of Robotics Research}, 31\penalty0
  (7):\penalty0 855--871, 2012.

\bibitem[Kolmogorov and Fomin(1957)]{kolmogorov1957}
Andrey~N Kolmogorov and Sergei~V Fomin.
\newblock \emph{{Elements of the Theory of Functions and Functional Analysis}}.
\newblock Dover, 1957.

\bibitem[Larsen et~al.(1999)Larsen, Gottschalk, Lin, and Manocha]{larsen1999}
Eric Larsen, Stefan Gottschalk, Ming~C Lin, and Dinesh Manocha.
\newblock {Fast proximity queries with swept sphere volumes}.
\newblock Technical report, TR99-018, Department of Computer Science,
  University of North Carolina, 1999.

\bibitem[Lin and Canny(1991)]{lin1991}
Ming~C Lin and John~F Canny.
\newblock {A fast algorithm for incremental distance calculation}.
\newblock In \emph{IEEE International Conference on Robotics and Automation
  (ICRA), 1991}, pages 1008--1014. IEEE, 1991.

\bibitem[Manocha and Demmel(1994)]{manocha1994}
Dinesh Manocha and James Demmel.
\newblock {Algorithms for intersecting parametric and algebraic curves I:
  Simple intersections}.
\newblock \emph{ACM Transactions on Graphics (TOG)}, 13\penalty0 (1):\penalty0
  73--100, 1994.

\bibitem[Mellinger et~al.(2012)Mellinger, Kushleyev, and Kumar]{mellinger2012}
Daniel Mellinger, Alex Kushleyev, and Vijay Kumar.
\newblock {Mixed-integer quadratic program trajectory generation for
  heterogeneous quadrotor teams}.
\newblock In \emph{Robotics and Automation (ICRA), 2012 IEEE International
  Conference on}, pages 477--483. IEEE, 2012.

\bibitem[Oh et~al.(2012)Oh, Kim, Lee, Kim, and Elber]{oh2012}
Young-Taek Oh, Yong-Joon Kim, Jieun Lee, Myung-Soo Kim, and Gershon Elber.
\newblock {Efficient point-projection to freeform curves and surfaces}.
\newblock \emph{Computer Aided Geometric Design}, 29\penalty0 (5):\penalty0
  242--254, 2012.

\bibitem[Pan et~al.(2012)Pan, Zhang, and Manocha]{pan2012}
Jia Pan, Liangjun Zhang, and Dinesh Manocha.
\newblock {Collision-free and smooth trajectory computation in cluttered
  environments}.
\newblock \emph{The International Journal of Robotics Research}, 31\penalty0
  (10):\penalty0 1155--1175, 2012.

\bibitem[Patterson et~al.(2019)Patterson, Lakshmanan, and
  Hovakimyan]{patterson2019}
Andrew Patterson, Arun Lakshmanan, and Naira Hovakimyan.
\newblock Intent-aware probabilistic trajectory estimation for collision
  prediction with uncertainty quantification.
\newblock \emph{arXiv preprint arXiv:1904.02765}, 2019.

\bibitem[Puig-Navarro et~al.(2018)Puig-Navarro, Hovakimyan, Alexandrov, and
  Allen]{puig2018}
Javier Puig-Navarro, Naira Hovakimyan, Natalia Alexandrov, and Bonnie~D Allen.
\newblock {Silhouette-informed trajectory generation through a wire maze for
  UAS}.
\newblock In \emph{2018 Aviation Technology, Integration, and Operations
  Conference}, page 3845, 2018.

\bibitem[Ricciardi and Vasile(2018)]{ricciardi2018}
Lorenzo~A Ricciardi and Massimiliano Vasile.
\newblock Direct transcription of optimal control problems with finite elements
  on {Bernstein} basis.
\newblock \emph{Journal of Guidance, Control, and Dynamics}, pages 1--15, 2018.

\bibitem[Rimon and Boyd(1997)]{rimon1997}
Elon Rimon and Stephen~P Boyd.
\newblock {Obstacle collision detection using best ellipsoid fit}.
\newblock \emph{Journal of Intelligent and Robotic Systems}, 18\penalty0
  (2):\penalty0 105--126, 1997.

\bibitem[Ross and Karpenko(2012)]{ross2012}
Isaac~M Ross and Mark Karpenko.
\newblock {A review of pseudospectral optimal control: From theory to flight}.
\newblock \emph{Annual Reviews in Control}, 36\penalty0 (2):\penalty0 182--197,
  2012.

\bibitem[Sederberg et~al.(1989)Sederberg, White, and Zundel]{sederberg1989}
Thomas~W Sederberg, Scott~C White, and Alan~K Zundel.
\newblock Fat arcs: A bounding region with cubic convergence.
\newblock \emph{Computer Aided Geometric Design}, 6\penalty0 (3):\penalty0
  205--218, 1989.

\bibitem[Tang et~al.(2009)Tang, Kim, and Manocha]{tang2009}
Min Tang, Young~J Kim, and Dinesh Manocha.
\newblock {{$C^2A$}: Controlled conservative advancement for continuous
  collision detection of polygonal models}.
\newblock In \emph{IEEE International Conference on Robotics and Automation
  (ICRA), 2009}, pages 849--854. IEEE, 2009.

\bibitem[Van~Loock et~al.(2015)Van~Loock, Pipeleers, and Swevers]{van2015}
Wannes Van~Loock, Goele Pipeleers, and Jan Swevers.
\newblock {Optimal motion planning for differentially flat systems with
  guaranteed constraint satisfaction}.
\newblock In \emph{American Control Conference (ACC), 2015}, pages 4245--4250.
  IEEE, 2015.

\bibitem[Váňa and Faigl(2018)]{vana2018}
Petr Váňa and Jan Faigl.
\newblock {Optimal solution of the generalized Dubins interval problem}.
\newblock In \emph{Proceedings of Robotics: Science and Systems}, Pittsburgh,
  Pennsylvania, June 2018.
\newblock \doi{10.15607/RSS.2018.XIV.035}.

\bibitem[Weisstein(2003{\natexlab{a}})]{weisstein2003fish}
Eric~W Weisstein.
\newblock {Fish Curve}.
\newblock 2003{\natexlab{a}}.

\bibitem[Weisstein(2003{\natexlab{b}})]{weisstein2003heart}
Eric~W Weisstein.
\newblock {Heart Curve}.
\newblock 2003{\natexlab{b}}.

\bibitem[Weisstein(2003{\natexlab{c}})]{weisstein2003ranunculoid}
Eric~W Weisstein.
\newblock {Ranunculoid}.
\newblock 2003{\natexlab{c}}.

\bibitem[Zhang et~al.(2006)Zhang, Lee, and Kim]{zhang2006}
Xinyu Zhang, Minkyoung Lee, and Young~J Kim.
\newblock {Interactive continuous collision detection for non-convex
  polyhedra}.
\newblock \emph{The Visual Computer}, 22\penalty0 (9-11):\penalty0 749--760,
  2006.

\end{thebibliography}

\def\black#1{{\color{black}#1}}
\def\red#1{{\color{red}#1}}
\def\blue#1{{\color{blue}#1}}
\def\green#1{{\color{green}#1}}
\def\yellow#1{{\color{yellow}#1}}

\def\Pan#1{\color{red}#1\normalcolor}
\def\PanCancel#1{\color{brown} \st{#1}\normalcolor}

\end{document}